\documentclass[acmsmall]{acmart}

\AtBeginDocument{%
  }

\setcopyright{acmcopyright}
\copyrightyear{2023}
\acmYear{2023}
\acmDOI{XXXXXXX.XXXXXXX}

\acmJournal{JACM}
\acmVolume{37}
\acmNumber{4}
\acmMonth{8}
\usepackage{subfigure}
\usepackage{textcomp}
\usepackage{stfloats}
\usepackage{url}
\usepackage{verbatim}
\usepackage{graphicx}
\usepackage{booktabs}
\usepackage{mathrsfs}
\usepackage{mathtools}
\usepackage{color}
\usepackage{xcolor}
\usepackage{arydshln}
\usepackage{colortbl}
\usepackage{mdframed}
\usepackage{soul}

\begin{document}

\title{Distinguish Confusion in Legal Judgment Prediction via Revised Relation Knowledge}

\author{Nuo Xu}
\email{nxu@sei.xjtu.edu.cn}
\orcid{1234-5678-9012}
\affiliation{%
  \institution{MOE KLINNS Lab, Xi’an Jiaotong University}
  \city{Xi'an}
  \country{China}
}

\author{Pinghui Wang}
\email{phwang@mail.xjtu.edu.cn}
\affiliation{%
  \institution{MOE KLINNS Lab, Xi’an Jiaotong University}
  \city{Xi'an}
  \country{China}
}

\author{Junzhou Zhao}
\email{junzhou.zhao@xjtu.edu.cn}
\affiliation{%
  \institution{MOE KLINNS Lab, Xi’an Jiaotong University}
  \city{Xi'an}
  \country{China}
}

\author{Feiyang Sun}
\email{fysun@sei.xjtu.edu.cn}
\affiliation{%
  \institution{MOE KLINNS Lab, Xi’an Jiaotong University}
  \city{Xi'an}
  \country{China}
}

\author{Lin Lan}
\email{llan@sei.xjtu.edu.cn}
\affiliation{%
 \institution{MOE KLINNS Lab, Xi’an Jiaotong University}
  \city{Xi'an}
  \country{China}
}

\author{Jing Tao}
\email{jtao@mail.xjtu.edu.cn}
\affiliation{%
  \institution{Zhejiang Research Institute, Xi’an Jiaotong University}
  \city{Hangzhou}
  \country{China;}
  \institution{MOE KLINNS Lab, Xi’an Jiaotong University}
  \city{Xi'an}
  \country{China}
}

\author{Li Pan}
\email{panli@sjtu.edu.cn}
\affiliation{%
  \institution{the department of electronic engineering, school of electronic information and electrical engineering, Shanghai Jiaotong University}
  \city{Shanghai}
  \country{China}
}

\author{Xiaohong Guan}
\email{xhguan@xjtu.edu.cn}
\affiliation{%
  \institution{MOE KLINNS Lab, Xi’an Jiaotong University}
  \city{Xi'an}
  \country{China;}
  \institution{Tsinghua National Lab for Information Science and Technology, Tsinghua University}
  \city{Beijing}
  \country{China}
  }

\renewcommand{\shortauthors}{Xu et al.}
\newcommand{\lzm}[1]{\textcolor{red}{LZ: #1}}
\begin{abstract}
Legal Judgment Prediction (LJP) aims to automatically predict a law case's judgment results based on the text description of its facts.
In practice, the confusing law articles (or charges) problem frequently occurs, reflecting that the law cases applicable to similar articles (or charges) tend to be misjudged.
Although some recent works based on prior knowledge solve this issue well, they ignore that confusion also occurs between law articles with a high posterior semantic similarity due to the data imbalance problem instead of only between the prior highly similar ones, which is this work's further finding.
This paper proposes an end-to-end model named \textit{D-LADAN} to solve the above challenges.
On the one hand, D-LADAN constructs a graph among law articles based on their text definition and proposes a graph distillation operation (GDO) to distinguish the ones with a high prior semantic similarity.
On the other hand, D-LADAN presents a novel momentum-updated memory mechanism to dynamically sense the posterior similarity between law articles (or charges) and a weighted GDO to adaptively capture the distinctions for revising the inductive bias caused by the data imbalance problem.
We perform extensive experiments to demonstrate that D-LADAN significantly outperforms state-of-the-art methods in accuracy and robustness.
\end{abstract}




\maketitle

\section{Introduction} \label{sec:introduction}
The application of artificial intelligence models to assist with legal judgment has become popular in recent years.
Legal judgment prediction (\textbf{LJP}) aims to predict a case's judgment results, such as applicable law articles, charges, and terms of penalty, based on its fact description, as illustrated in Table~\ref{tab:LJP_task}.
Such a technique can not only assist judiciary workers in processing cases but also offer legal consultancy services to the public.
Previous literature typically formulates the LJP as a joint task with three text classification subtasks: applicable law article prediction, charge prediction, and the term of penalty prediction.
Multifarious methods have been proposed and got some successes, from the early rule-based methods~\cite{liu2004case,lin2012exploiting} to the recent neural-based models~\cite{hu2018few,luo2017learning,zhong2018legal,xu2020Ladan,gan2021Logic,dong2021R_former}.


\begin{table}[t]
\caption{An illustration of the LJP in the civil law system. 
        Generally, a judge needs to conduct professional analysis and reasoning on the fact description of the case and then choose applicable law articles, charges, and the term of penalty to convict the offender.}
\label{tab:LJP_task}
\centering
\begin{tabular}{l}
\toprule
\textbf{Fact Description}\\
\parbox[l]{0.95\columnwidth}{From Jan. 2006 to Mar. 2007, the defendant Gong \textcolor{red}{\textbf{used the position of signing and clearing the contract}} for the coal unloading business of the plan’s steam coal during \textcolor{blue}{\underline{\textit{his service as a company manager}}} to obtain benefits for a loading and unloading team. And \textcolor{red}{\textbf{he illegally accepted the benefit fee of 40,000 yuan}} from the legal person of the loading and unloading team...}\\
\midrule
\textbf{Relevant Law Article}\\
Criminal Law of the People’s Republic of China\\
\parbox[l]{0.95\columnwidth}{\textbf{Article 163:} \textcolor{green!50!black}{\textbf{[Bribery crime of non-state staffs]}} \textcolor{blue}{\underline{\textit{The employees of companies,}}}  enterprises or other units who, \textcolor{red}{\textbf{taking advantage of his position}}, demands money or property from another person, or \textcolor{red}{\textbf{illegally accepts another person's money}} or property in return for securing benefits for the person, if the amount involved is relatively large, shall be sentenced to \textcolor{blue!60!cyan}{\textbf{fixed-term imprisonment of not more than three years}} or criminal detention and shall also be fined...}\\
\midrule
\textbf{Charge: \textcolor{green!50!black}{Bribery crime of non-state staffs}}\\
\midrule
\textbf{Term of Penalty: \textcolor{blue!60!cyan}{A fixed-term imprisonment of twenty-four months}}\\
\bottomrule
\end{tabular}
\end{table}

\begin{figure}[t]
\centering
\includegraphics[width=1.0\linewidth]{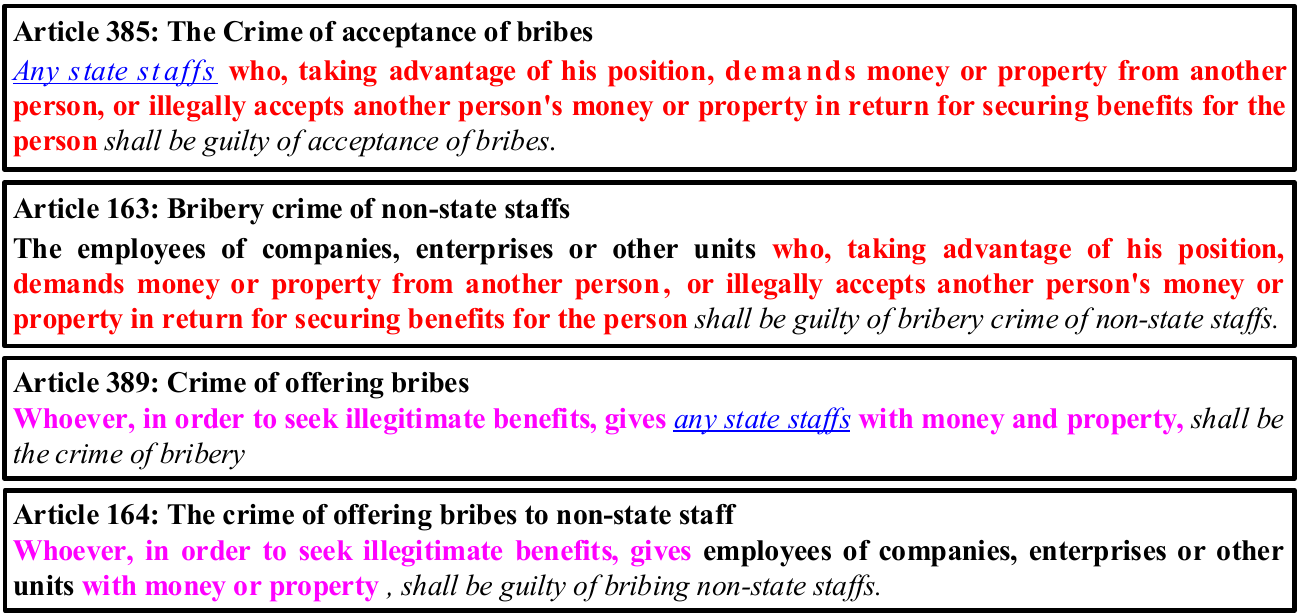}
\caption{Examples of prior confusing charges. We marked similar text with the same pattern, such as \textbf{bold}, \textit{italic}, red font, pink font, and blue underline.}
\label{fig:law_artices}
\end{figure}

A main drawback of existing methods is that they fail to solve the issue of {\em confusing law articles (or charges)}.
This issue describes the situation where similar cases tend to be misjudged against each other due to the high similarity of their corresponding law articles (or charges).
For example, in Fig.~\ref{fig:law_artices}, \textit{Article 385} and
\textit{Article 163} all describe offenses of accepting bribes, and their subtle differences are whether the guilty parties are state staff.
From this perspective, the key challenge to solving the confusing charges issue is to capture essential but rare features for distinguishing confusing law articles.

\begin{figure}[t]
\centering
\includegraphics[width=0.85\linewidth]{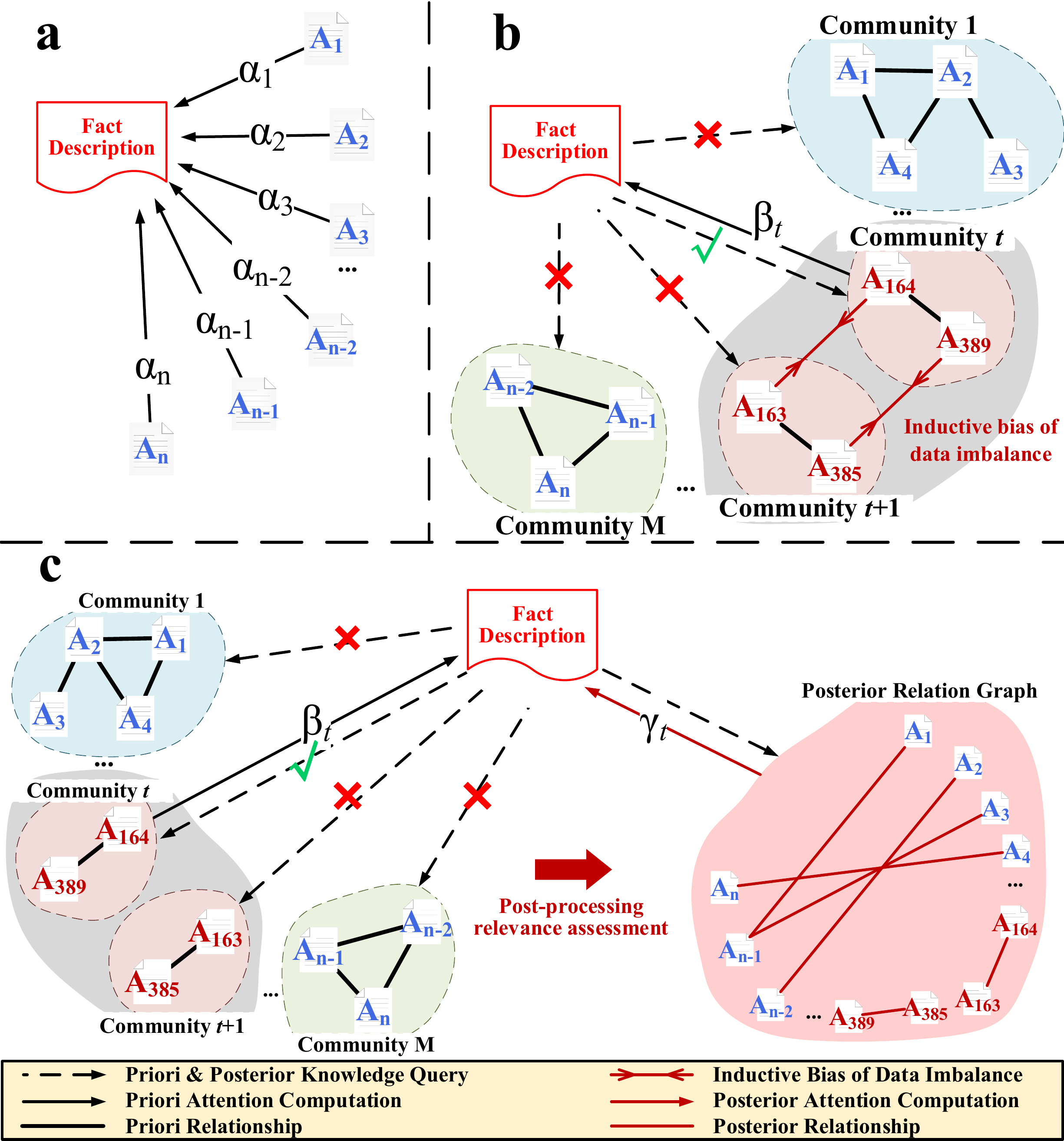}
\caption{\textbf{a.}~The fact-law attention framework of ~\cite{luo2017learning}.
  \textbf{b}.~The attention framework of LADAN, where the inductive bias caused by the data imbalance problem would damage the prior relationship structure.
  \textbf{c}.~Our framework further revises the inductive bias of the data imbalance by constructing a posterior relational structure. 
  \textit{Variables $\mathbf{\alpha}$, $\mathbf{\beta}$ and $\mathbf{\gamma}$ represent the context vectors learned from law articles for attentively extracting features from fact descriptions.}} \label{fig:attention_frameworks}
\end{figure}

To distinguish confusing charges to solve this issue, Hu et al.~\cite{hu2018few} define ten discriminative attributes, and Lyu et al.~\cite{lyu2022CEEN} propose four types of criminal elements.
However, the heavy dependence on expert knowledge of the definition and labeling makes applying these two methods to different civil law systems hard.
Inspired by the fact that Luo et al.~\cite{luo2017learning} use the attention vector of law articles to extract the corresponding key feature from fact description, a recent important direction of recent research to solve the confusing law articles (or charges) is to mine valid information from prior knowledge of law articles and charges for extracting distinguishable semantic features from the fact descriptions.
Different from the framework of~\cite{luo2017learning}(as shown in Fig.~\ref{fig:attention_frameworks}a), where each law independently attentively extracts features from fact description, recent works generally model the prior relationship between law articles (or charges) to capture the distinguishable features.
The most typical of them is our previous version, i.e., LADAN~\cite{xu2020Ladan}, whose framework is in Fig.~\ref{fig:attention_frameworks}b.
LADAN constructs the graph structure of law articles based on the prior word frequency similarity and divides law articles into several communities, in each of which law articles are easy to be confused from the perspective of prior. 
Then, it proposes a graph distillation operator to learn the differences between confusing law articles and attentively extract the distinguishable features from cases' fact descriptions. 
The subsequent studies more or less follow this insight to solve the confusing law article (or charge) issue. 
For example, Dong et al.~\cite{dong2021R_former} further model the 'article-charge-penalty' relationship and solve LJP as a node classification task by proposing a graph reasoning network.
Zhang et al.~\cite{zhang2023CL4LJP} then extend these prior relationships to the instance level and present a contrastive learning framework to solve JLP. 
They consider cases whose applicable law articles or charges with co-ownership as negative samples and achieve state-of-the-art performance.

\begin{figure}[t]
  \centering
  \subfigure[\label{fig:Law_topjudge}]{\includegraphics[width=0.45\linewidth]{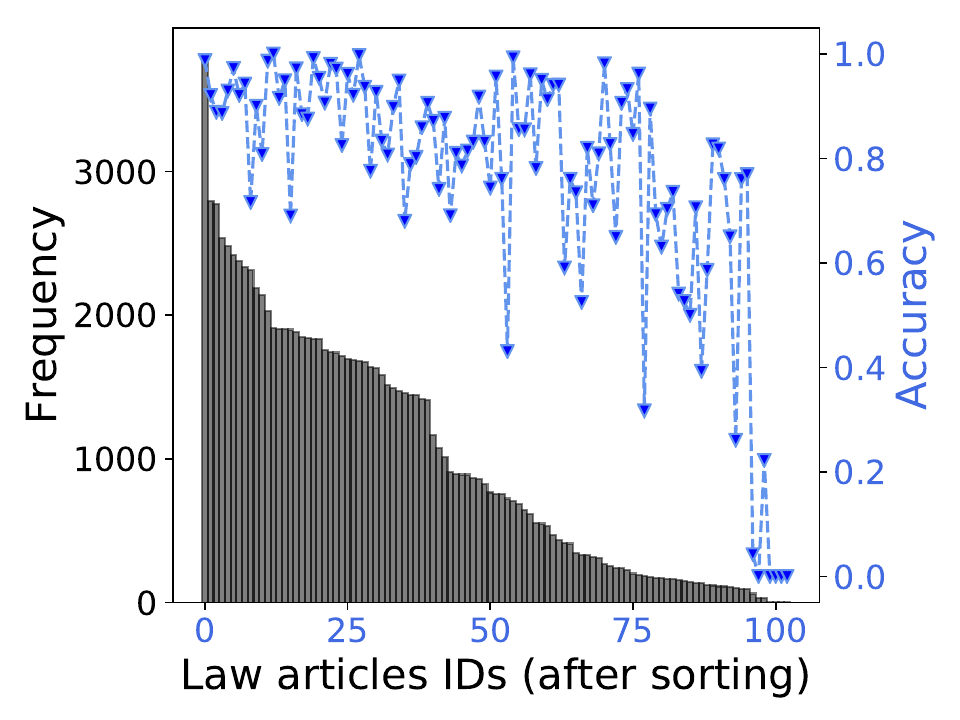}}
  \hspace{5mm}
  \subfigure[\label{fig:Accu_topjudge}]{\includegraphics[width=0.45\linewidth]{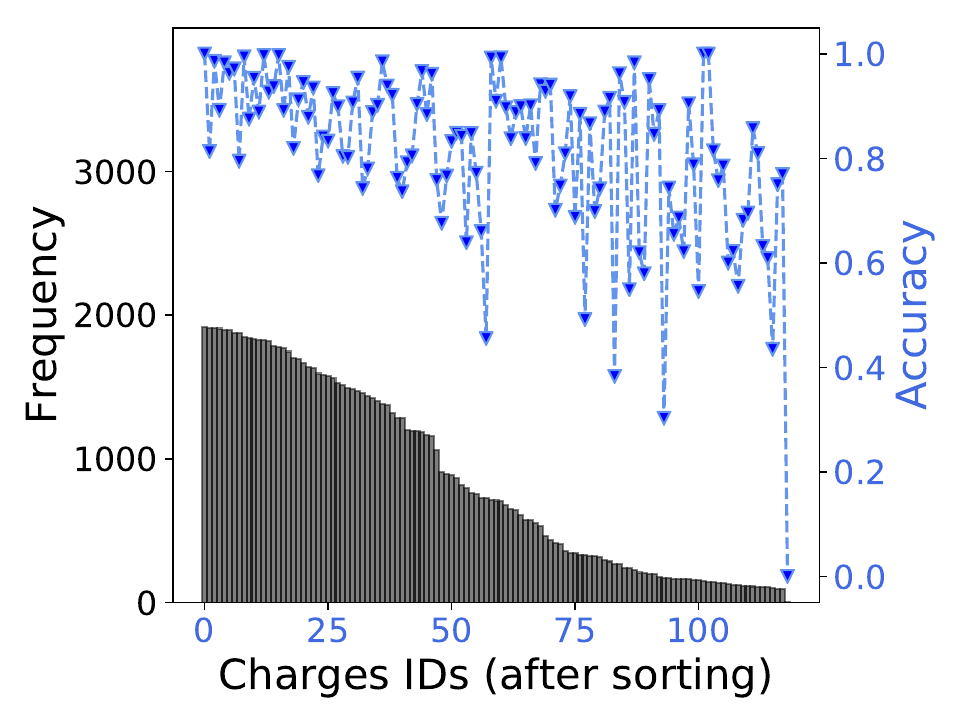}}
\caption{The frequency distribution of the law articles and charges on CAIL-small and the corresponding accuracy of LADAN on each category. Note that the IDs of the x-axis have been sorted in descending order of frequency.}
\label{fig:Acc_frequency}
\end{figure}

Although these works have made full use of the prior knowledge of the law and achieved considerable results, Zhang et al.~\cite{zhang2023CL4LJP} still state that existing methods are not ideal for the improvement of the tail category. As evidence, Fig.~\ref{fig:Acc_frequency} shows the LADAN's accuracy on different frequency law articles and charges.
We see that the LJP task is under a data imbalance distribution, and the performance of LADAN decreases with the decrease in frequency.
We argue that the inductive bias caused by the data imbalance problem destroys the pre-established relationship structure based on prior knowledge and causes posterior confusing law articles (or charges). The data imbalance problem reflects a phenomenon that the many-shot knowledge (or elements) covers the few-shot one during the learning process.
For ease of understanding, we take the following example that combines Fig.~\ref{fig:law_artices} and~\ref{fig:attention_frameworks}b.
\begin{example}{
To distinguish the four similar confusing law articles (i.e., \textit{Article 163, 164, 385, and 389} in Fig.~\ref{fig:law_artices}), LADAN divides them into two groups (i.e., \textit{Community $t$ and $t+1$} in Fig.~\ref{fig:attention_frameworks}b).
Such a framework tries to identify the subtle differences between \textit{Article 163} and \textit{389} (whether parties are state staffs), which premise that the visible difference between \textit{Community $t$} and \textit{$t+1$} (giving or receiving the property) are easy to learn.
However, when there are far fewer legal cases for giving property than ones for accepting property, the model is subject to inductive bias from unbalanced data and will misjudge all cases with giving property as the ones with receiving property due to similar contexts.
Thus, the above significant prior differences turn into difficulties to distinguish for LADAN.
In other words, in model understanding, the graph structure between law articles will have two more edges than the prior one (i.e., the red edges in Fig.~\ref{fig:attention_frameworks}b), which destroys LADAN's premise assumption and leads to model recognition confusion failure.
Similar examples can extend to other prior-relation-based models.
}
\end{example}\label{exp:1.1}

We refer to the issue that the above examples reflect in this paper as \textbf{posterior confusion} induced by data imbalance.
Since the above example may blur the distinction between posterior confusion and the few-shot charge problem introduced by Hu et al.~\cite{hu2018few}, we clarify the difference here to avoid misunderstanding.
Different from the few-shot charge problem that only considers data imbalance at the category level,
the posterior confusion considers the imbalance distribution among some more fine-grained key elements (e.g., whether the defendant accepts or gives property in \emph{Example 1.\ref{exp:1.1}}).
So posterior confusion also impairs the performance of some middle categories whose essential key elements are a small percentage of the total training samples, instead of only the tail categories.
Also using \emph{Example 1.\ref{exp:1.1}} as an example, 
if \textit{Article 164} is the top category, \textit{Article 389} and \textit{385} are middle categories, and \textit{Article 163} is the tail category, the posterior confusion issue would also make the article confusing occur between \textit{Article 389} and \textit{385}. 
This is because the number of the top category samples is too large, causing the model to blur the difference between giving and receiving property, treating the former as the latter. 
In a word, posterior confusion is a more complex and detailed issue.

To solve the confusing law articles problem, we propose an end-to-end framework, i.e.,
\textbf{D}ynamic \textbf{L}aw \textbf{A}rticle \textbf{D}istillation based \textbf{A}ttention \textbf{N}etwork (\textbf{D-LADAN}), whose framework is in Fig.~\ref{fig:attention_frameworks}c. 
D-LADAN takes the way of post-processing perception to correct the damage caused by the inductive bias of the data imbalance problem to the prior relationship of law articles. To be more specific:

\textbf{On the one hand}, to fully use the prior knowledge of the law articles, D-LADAN follows the assumption of its previous version LADAN: it is easy to distinguish dissimilar law articles as sufficient distinctions exist but challenging to recognize between similar law articles due to the few useful features.
D-LADAN first groups law articles into different communities, in each of which the law articles are highly semantically similar.
Then we propose a graph-based representation learning method to automatically explore the differences among law articles and compute an attention vector for each community.
For an input law case, we learn both macro and micro-level features.
D-LADAN uses macro-level features to predict the community containing the corresponding applicable law articles.
Then, it extracts micro-level features attentively by the attention vector of the selected community for distinguishing confusing law articles within the same community.

\textbf{On the other hand}, to solve the posterior confusion problem and make the model correctly learn the similarity relation of law articles, D-LADAN follows a straightforward but effective idea: if the data imbalance problem does indeed destroy the prior relationship of law articles, then its external manifestation is causing a new confusion, i.e., relationships with a posterior high semantic similarity.
Thus, D-LADAN uses a revised memory to dynamically capture the posterior similarity relation that the model learned in the training process and extracts the distinguishable features of fact descriptions to correct the inductive bias caused by the data imbalance problem from a posterior perspective.
In addition, D-LADAN uses a momentum updating mechanism to ensure the end-to-end training character of the model. 
Combining the prior and the revised similarity relation knowledge, the model can further learn more correct relations of law articles and improve its performance.

Our main contributions are summarized as follows:
\begin{enumerate}
\item{We develop an end-to-end framework, D-LADAN, to solve the LJP task.
To the best of our knowledge, it's the first work to discuss the posterior confusion issues from the perspective of the data imbalance problem.}

\item{We propose a novel graph distillation operator (GDO) with its weighted version to extract distinguishable features and effectively distinguish confusing law articles.}

\item{We propose a momentum-updated memory mechanism to capture the similarity relation of categories that the model learned and revise the inductive bias caused by the data imbalance problem. This module further improves the performance effectively.}

\item{We conduct extensive experiments on five real-world datasets.
The results show that our model outperforms state-of-the-art methods. To facilitate future research, we make our code publicly available\footnote{Our source codes are available at \url{https://github.com/prometheusXN/D-LADAN}.}.}
\end{enumerate}

We organize the rest of this paper as follows. 
Section~\ref{sec:related} summarizes related work.
Section~\ref{sec:problem} formulates the problem.
Section~\ref{sec:method} presents our method D-LADAN.
The performance evaluation and testing results are in
Section~\ref{sec:experiment}.
Conclusion remarks then follow.  
\section{Related Work} \label{sec:related}
Our work solves the problem of the confusing charge in the LJP task by referring to the calculation principle of the graph neural networks (GNNs).
In this section, we will introduce related works from these two aspects.

\subsection{Legal Judgment Prediction}
The early works on LJP focus on analyzing existing legal cases in specific scenarios with mathematical and statistical algorithms~\cite{kort1957predicting,nagel1963,keown1980mathematical,lauderdale2012supreme}. 
Besides, some studies developed machine learning-based methods \cite{lin2012exploiting,liu2004case,sulea2017exploring} to solve the problem of LJP, which almost combines some manually designed features with a linear classifier to improve the performance of case classification.
The shortcoming is that these methods rely heavily on manual feature engineering and suffer from the generalization problem.

In recent years, due to the rapid development of neural networks, researchers have widely used neural networks to solve LJP tasks, mainly divided into two main lines of work.
The first line of work focuses on improving the performance by investigating the relationship between the three subtasks, i.e., the relevant law article prediction, charge prediction, and term of penalty prediction.
Zhong et al.~\cite{zhong2018legal} first model the explicit dependencies among subtasks with scalable directed acyclic graph forms and propose a topological multi-task learning framework for effectively solving these subtasks together.
Yang et al.~\cite{yang2019legal}, inspired by judges' reconfirming behavior after the evaluation, refined this framework by adding backward dependencies between the prediction results of subtasks.
Further considering the entangled relationships of labels inter- and intra-subtasks, Dong and Niu~\cite{dong2021R_former} organize all law articles, charges, and the term of penalty labels into a large graph and formalize LJP as a graph node classification task to solve the problem.
Yue et al.~\cite{yue2021Neurjudge} further deepen the study of the relationship between subtasks and legal case instances.
They decouple the fact description into the three corresponding conditions for each subtask and create predictions, where the intermediate results also obey the forward dependencies of subtasks at the same time.

The second line of work is looking at how best to use the semantic information of laws and charges to help solve the LJP task.
Luo et al.~\cite{luo2017learning} propose a hierarchical attentional network to capture the relation between fact description and relevant law articles to improve the charge prediction.
To improve the interpretability of their model, Gan et al.~\cite{gan2021Logic} first transform declarative legal knowledge into logic rules and then constrain their model with these rules.
To the best of our knowledge, Hu et al.~\cite{hu2018few} are the first to study the problem of discriminating confusing charges for automatically predicting applicable charges.
They manually define ten discriminative attributes and propose to enhance the representation of the case fact description by learning these attributes.
Then, Lyu et al.~\cite{lyu2022CEEN} proposed four types of criminal elements to distinguish confusing law articles and recognize fairly similar fact descriptions.
As these methods rely heavily on experts and cannot be extended to different law systems easily, some works try to automatically extract crucial attributes to solve the confusing problem by proposing novel model frameworks. 
The earlier conference version of our work~\cite{xu2020Ladan} uses GNNs to automatically derive the differences between confusing legal articles to improve the representation of fact description.
Extending the differences between law articles and charge labels to the instance level, Zhang et al.~\cite{zhang2023CL4LJP} propose a contrastive learning framework with three learning tasks to assist the model in differentiating between similar law articles and charges, which achieves state-of-the-art performance.

However, these frameworks ignore the imbalance problem, which leads the model to learn biased similarity relationships between labels (i.e., law articles and charges).
To solve this issue, we modify the previous version of LADAN with a momentum-updated revised memory mechanism, which dynamically captures the semantic similarity relationship between law articles the model learned. By adaptively distilling the differences from the revised memory, the representations of fact description get further enhanced.

\subsection{Graph Neural Networks}
Due to their excellent performance in graph structure data, GNNs have attracted significant attention~\cite{kipf2016semi, hamilton2017inductive, DSE2019exploring} recently.
In general, existing GNNs focus on proposing different aggregation schemes to fuse features from the neighborhood of each node in the graph for extracting richer and more comprehensive information.
Kipf et al.~\cite{kipf2016semi} propose a graph convolution network that uses mean pooling to pool neighborhood information.
GraphSAGE~\cite{hamilton2017inductive} concatenates the node's features and applies the mean/max/LSTM operators to aggregate neighborhood information for inductively learning node embeddings.
MR-GNN~\cite{xu2019mr} aggregates the multi-resolution features of each node to exploit node information, subgraph information, and global information together.
Besides, Message Passing Neural Networks~\cite{gilmer2017neural} further consider edge information when they are doing the aggregation.
However, the aggregation schemes lead to the over-squashing and over-smoothing issues of GNNs~\cite{li2018deeper,topping2021over_squashing, Tackling_oversmoothing}, i.e., the aggregated node representations would become indistinguishable, which is entirely contrary to our goal of extracting distinguishable information.
Therefore, we propose our graph distillation operation based on a distillation strategy instead of aggregation schemes to capture the distinguishable features between similar law articles.

\section{Problem Formulation} \label{sec:problem}
\begin{table}[t]
\caption{Main mathematical notations.}\label{tab:Notation}
\centering
\begin{tabular}{c|c}
\toprule 
Notation & Description\\
\midrule 
$f = \{S_1, \cdots, S_{n_f}\}$ & the sentence sequence of a fact description\\
$S_i = \{w_{i,1}, \cdots, w_{i,n_i}\}$ & the word sequence of the sentence $S_i$\\
$\mathcal{L}=\{L_1,\ldots,L_{m}\}$ & the set of law articles\\
$\mathcal{M} = \{\mathbf{m}_{L_1}, \cdots, \mathbf{m}_{L_m}\}$ & the set of revised memory vector representations of law articles\\
$G = \{g_1, \cdots, g_k\}$ & the prior graph, which is a set of law article communities (subgraphs)\\
$G_M = \{\mathcal{M}, \mathcal{A}_{M}\}$ & the full-connected graph of the revised memory $\mathcal{M}$ \\
$Y_l = \{y_{l,1}, \cdots, y_{l, \lvert Y_{l}\lvert}\}$ & the set of law article labels\\
$Y_c = \{y_{c,1}, \cdots, y_{c, \lvert Y_{c}\lvert}\}$ & the set of charge labels\\
$Y_t = \{y_{t,1}, \cdots, y_{t, \lvert Y_{t}\lvert}\}$ & the set of term of penalty labels\\
$\mathbf{w}_{i,j} $ & the word embedding of word $w_{i,j}$\\
$\mathbf{v}_{f}, \mathbf{v}_{S_i}, \mathbf{v}_{L_i} $ & the vector representation of the corresponding $f$, $S_i$, and $L_i$\\

\bottomrule
\end{tabular}
\end{table}
In this section, we introduce some notations and terminologies. Then we formulate the LJP task. 
To facilitate understanding of the subsequent method introductions, we summarize the main notations in Table~\ref{tab:Notation}.

\noindent {\bf{Law Cases.}}
Each law case consists of a \textit{fact description} and several \textit{judgment results} (cf.~Table.~\ref{tab:LJP_task}).
The fact description is represented as a text document, denoted by $f$.
The judgment results include \textit{applicable law articles}, \textit{charges}, \textit{terms of penalty}, whose label sets are denoted by $Y_l$, $Y_c$, and $Y_t$ respectively.
We use $\lvert Y_{*}\lvert$ to represent the number of labels for the corresponding judgment result. 
Thus, a law case can be represented by a tuple $(f, y_l, y_c, y_t)$.

\noindent {\bf{Law Articles.}}
Law cases are often analyzed and adjudicated according to a legislature's \textit{statutory law} (also known as \textit{written law}).
Formally, we denote the statutory law as a set of \textit{law articles} $\mathcal{L}=\{L_1, \ldots, L_{m}\}$, where $m$ is the number of law articles.
Similar to the fact description of cases, we also represent each law article $L_i$ as a document.

\noindent {\bf{Legal Judgment Prediction.}}
Given a training dataset $D = \{(f, y_l, y_c, y_t)_z\}_{z=1}^{q}$ of size $q$,
we aim to train a model $\text{F}(\cdot)$ that can predict the judgment results for any test law case with a fact description $f_{test}$, i.e., $\text{F}(f_{test}, \mathcal{L}) = (\hat{y}_l, \hat{y}_c, \hat{y}_t)$, where $\hat{y}_l$, $\hat{y}_c$, and $\hat{y}_t$ represent the predicted relevant law article, charge, and the term of penalty, respectively.
Following~\cite{zhong2018legal,yang2019legal}, we assume each case has only one applicable law article and charge.

\section{Our Method} \label{sec:method}
\subsection{Overview of Framework}\label{sec:overview}
\begin{figure*}[t]
\centering
\includegraphics[width=1.0\linewidth]{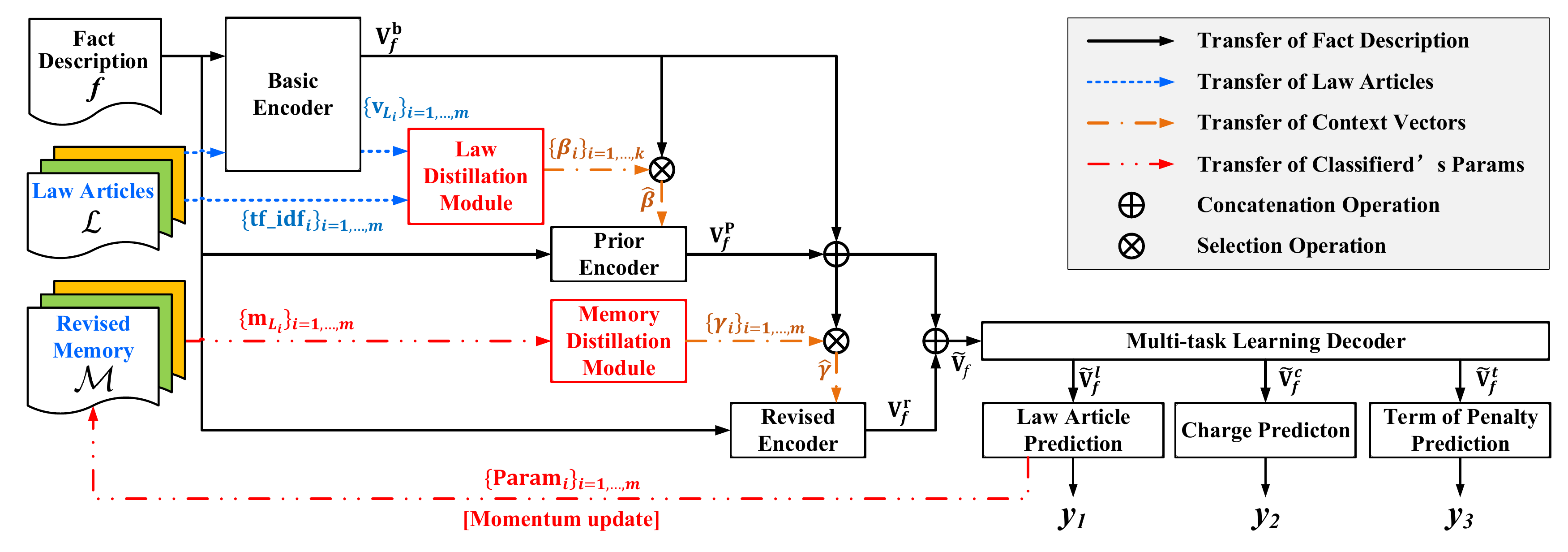}
\caption{Overview of our framework \textit{D-LADAN}: it takes the fact descriptions of cases and the text definitions of law articles as inputs. Then, it extracts the basic representation $\mathbf{v}_f^{\text{b}}$, the prior distinguishing representation $\mathbf{v}_f^{\text{p}}$, and the revised distinguishing representation $\mathbf{v}_f^{\text{r}}$ of the fact descriptions through the corresponding encoders. Finally, it combines these three representations for the downstream prediction tasks.}
\label{fig:framework}
\end{figure*}

In our framework D-LADAN (cf.~Fig.~\ref{fig:framework}), the representation of fact description of a case consists of three parts: 
a \emph{basic representation} denoted by $\mathbf{v}_f^{\text{b}}$, a prior distinguishable representation (\emph{regard as prior representation}) denoted by $\mathbf{v}_f^{\text{p}}$, and a revised distinguishable representation (\emph{regard as revised representation}) denoted by $\mathbf{v}_f^{\text{r}}$, i.e., $\tilde{\mathbf{v}}_f = [\mathbf{v}_f^{\text{b}} \oplus \mathbf{v}_f^{\text{p}} \oplus \mathbf{v}_f^{\text{r}}]$, where the symbol $\oplus$ denotes the concatenation operation.
The basic representation $\mathbf{v}_f^{\text{b}}$ contains basic semantic information for matching a group of law articles that may apply to the case.
In contrast, the prior representation $\mathbf{v}_f^{\text{p}}$ considers the semantic similarity relation between the prior definitions of law articles and captures features that can effectively distinguish confusing law articles.
In addition, to overcome the imbalance problem, the revised representation $\mathbf{v}_f^{\text{r}}$ dynamically senses the similarity relation of law articles learned by the model and adaptively mines the differences that further enhance the distinguishability of the vector representation of the description.
To more graphically illustrate how our D-LADAN solves the confusing problems mentioned above, we refer to Table~\ref{tab:LJP_task} and Fig.~\ref{fig:law_artices} and provide a practical example of what we expect D-LADAN to do, as follows,

\begin{example}
When getting the case's fact description of Table~\ref{tab:LJP_task} as input, D-LADAN encodes the full text equally into the basic representation $\mathbf{v}_f^{\text{b}}$.
Then, D-LADAN notices that \textit{Article 385} and \textit{163} are highly similar, it uses the \emph{law distillation module} to capture the difference between these two articles, i.e., the word "\textit{Any state staffs}" and "\textit{The employees of companies, enterprises or other units}".
D-LADAN's \textit{prior encoder} further uses such distinguishable keywords to focus attention on capturing the corresponding information from the fact description (i.e., "\textit{his service as a company manager}" words in Table~\ref{tab:LJP_task}) to get the prior representation $\mathbf{v}_f^{\text{p}}$.
During the model training, if D-LADAN finds that the representations of two law articles gradually become similar due to data imbalance (e.g., \textit{Article 163} and \textit{164}), it uses the \textit{memory distillation module} to capture the difference (e.g, the words "\textit{accept}" and "\textit{give}") between laws with a posterior high similarity.
D-LADAN's \textit{revised encoder} uses the posterior differences to focus attention on capturing the distinguishable information (i.e., "\textit{he illegally accepted the benefit fee of 40,000 yuan}" words in Table~\ref{tab:LJP_task}) to get the revised representation $\mathbf{v}_f^{\text{r}}$.
Combining these three representations, D-LADAN can solve the confusing problem in the LJP task well.
\end{example}

As we mentioned, it is easy to distinguish dissimilar law articles as sufficient distinctions exist, and the difficulty in solving confusing charges lies in extracting distinguishable features of similar law articles.
To obtain the basic representation $\mathbf{v}_f^{\text{b}}$, we choose a popular document encoding method (e.g., CNN encoder \cite{kim2014convolutional} or Bi-RNN encoder \cite{yang2016hierarchical}).
To learn the prior representation $\mathbf{v}_f^{\text{p}}$, we use the \emph{law distillation module} first to divide law articles into several communities to ensure that the law articles of each community are highly similar, and then extract each community $i$'s prior distinction vector (or prior distinguishable features) $\mathbf{\beta}_i$ from the basic representations of law articles in the community $i$.
Given the case's fact description, based on all communities' distinction vectors, we generate the most relevant one (i.e., $\mathbf{\hat{\beta}}$ in Fig.~\ref{fig:framework}) for attentively extracting the prior distinguishable features $\mathbf{v}_f^{\text{p}}$ in the subsequent fact re-encode module (i.e., the prior encoder in Fig.~\ref{fig:framework}).
To generate the revised representation $\mathbf{v}_f^{\text{r}}$, we propose a revised memory mechanism with the \emph{memory distillation module}, which considers the fully-connected similarity graph of memories and computes the revised distinction vector $\gamma_i$ for each memory $\mathbf{m}_i$. 
Then, we generate the most relevant one (i.e., $\mathbf{\hat{\gamma}}$ in Fig.~\ref{fig:framework}) for each case's fact description to capture the revised distinguishable feature by the similar fact re-encode module (i.e., the revised encoder in Fig.~\ref{fig:framework}).
In addition, in the training processing, we momentum update the revised memories with the parameters of the law article classifier after each training step to sense the semantic similarity relation between law articles that the model learned.
In the following, we will elaborate on the law distillation module (Sec.~\ref{sec:law_interaction}), the memory distillation module (Sec.~\ref{sec:memory_interaction}), and the fact re-encode module, i.e., the prior encoder and the revised encoder in Fig.~\ref{fig:framework}(Sec.~\ref{sec:re_encoder}).

\subsection{Distilling Law Articles} \label{sec:law_interaction}
\begin{figure*}[t] 
\centering
\includegraphics[width=1.0\linewidth]{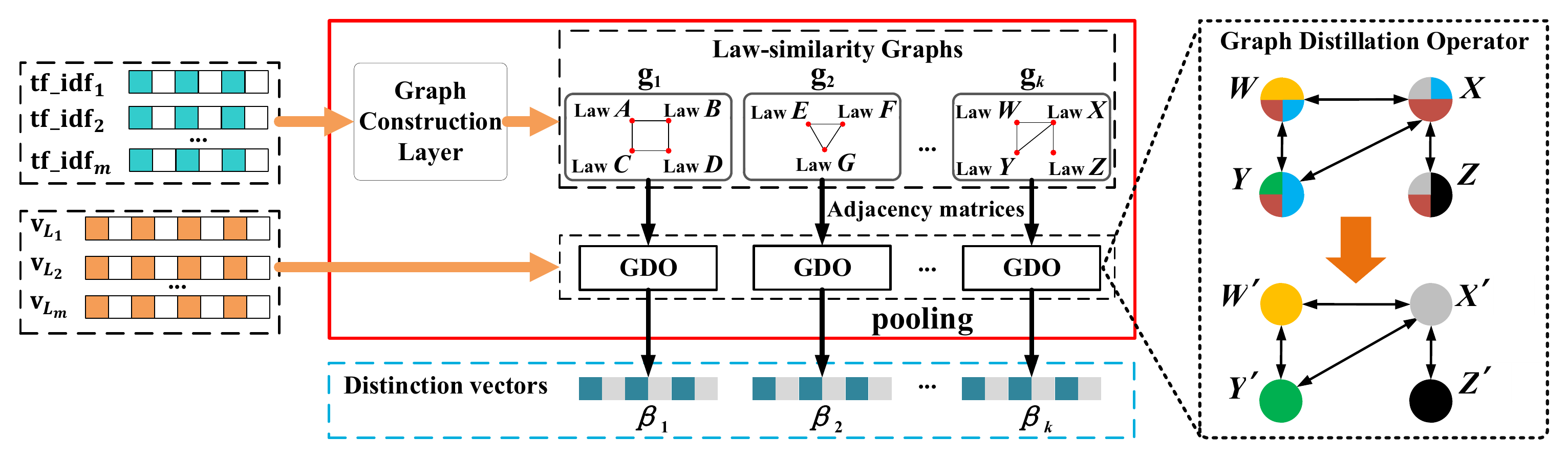}
\caption{Law Distillation Module: this module groups law articles based on the prior similarity relation and distills the distinguishable features of each community for attention calculation of the prior encoder.
}
\label{fig:Prior_GDO}
\end{figure*}

As mentioned earlier, a case might be misjudged due to the high similarity of some law articles.
To alleviate this problem, we design a law distillation module (cf. Fig.~\ref{fig:Prior_GDO}) to extract distinguishable and representative information from the prior definition of all law articles.
Specifically, it first uses a \emph{graph construction layer} (\textbf{GCL}) to divide law articles into different communities.
Then, we apply the graph distillation layer to learn the discriminative representation of each law article community, called the \textit{prior distinction vector} in the remainder of this paper.

\subsubsection{Graph Construction Layer} \label{sec:graph_prior}
To find probably confusing law articles, we first construct a fully-connected graph $G^*$ for all law articles $\mathcal{L}$,
where the weight on the edge between a pair of law articles $L_i, L_j\in \mathcal{L}$ is defined as the cosine similarity between the two articles' TF-IDF (Term Frequency-Inverse Document Frequency) representations $\mathbf{tf\_idf}_i$ and $\mathbf{tf\_idf}_j$.
Since confusing law articles are usually semantically similar
and there exists sufficient information to distinguish dissimilar law articles, we remove the edges with weights less than a predefined threshold $\theta$ from graph $G^*$.
By setting an appropriate threshold $\theta$, we obtain a new graph $G = \{g_i\}_{i = 1}^{k}$ composed of several disconnected subgraphs $g_1, \ldots, g_k$ (or, communities),
where each $g_i, i=1,\ldots, k$ contains a community of probably confusing articles.
Our later experimental results demonstrate that this easy-to-implement method
effectively improves the performance of D-LADAN.
\subsubsection{Graph Distillation Layer} \label{subsec:GDL}
To extract distinguishable information from each community $g_i$, a straightforward way is to delete duplicate words and sentences presented in law articles within the community (as described in Sec.~\ref{sec:introduction}).
In addition to introducing significant errors, this simple method cannot be plugged into end-to-end neural architectures due to its non-differentiability.
To overcome the above issues, inspired by the popular graph convolution operator (\textbf{GCO})~\cite{kipf2016semi,hamilton2017inductive,velivckovic2017graph}, we propose a \emph{graph distillation operator} (\textbf{GDO}) for effectively extracting distinguishable features.
In contrast to the GCO that computes the message propagation between neighbors and aggregates these messages to enrich representations of nodes in the graph, the basic idea behind our GDO is to learn efficient information with distinction by removing similar features between nodes.

Specifically, for an arbitrary law article $L_i$, GDO uses a trainable weight matrix $\Psi$ to capture similar information between it and its neighbors in the graph $G$, and a matrix $\Phi$ to extract effective semantic features of $L_i$.
At each layer $l\ge 0$, the aggregation of similar information between $L_i$ and its neighbors is removed from its representation, that is,
\[
 \mathbf{v}_{L_i}^{(l+1)} = \Phi_{\text{L}}^{(l)}\mathbf{v}_{L_i}^{(l)} - \sum_{L_j\in N_i}\frac{\Psi_{\text{L}}^{(l)}(\mathbf{v}_{L_i}^{(l)} \oplus \mathbf{v}_{L_j}^{(l)})}{|N_i|} + \mathbf{b}_{\text{L}}^{(l)},
\]
where $\mathbf{v}_{L_i}^{(l)} \in \mathbb{R}^{d_l}$ refers to the representation of law $L_i$ in the $l^\text{th}$ graph distillation layer, $N_i$ refers to the neighbor set of $L_i$ in graph $G$, $\mathbf{b}_{\text{L}}^{(l)}$ is the bias, and $\Phi_\text{L}^{(l)} \in \mathbb{R}^{d_{l+1} \times d_{l}}$ and $\Psi_\text{L}^{(l)} \in \mathbb{R}^{d_{l+1} \times 2d_{l}}$ are the trainable self-weighted matrix and the neighbor similarity extracting matrix for the law distillation module respectively. Note that $d_l$ is the dimension of the feature vector in the $l$-th graph distillation layer.
We set $d_0 = d_s$, where ${d_s}$ is the dimension of basic representations $\mathbf{v}_f^{\text{b}}$ and $\mathbf{v}_{L_i}$. Similar to GCO, our GDO also supports multi-layer stacking.

Using GDO with $H$ layers, we output law article representation of the last layer, i.e., $\mathbf{v}_{L_i}^{(H)} \in \mathbb{R}^{d_H}$, which contains rich distinguishable features that can distinguish law article $L_i$ from the articles within the same community.
To further improve law articles' distinguishable features,
for each subgraph $g_i, i = 1,2, \dots, k$ in graph $G$,
we compute its prior distinction vector $\beta_i$
by using pooling operators to aggregate the distinguishable features of articles in $g_i$. Formally, $\beta_i$ is computed as:
\[
 \beta_i = [\text{MaP}(\{\mathbf{v}_{L_i}^{(H)}\}_{L_j\in g_i}), \text{MiP}(\{\mathbf{v}_{L_i}^{(H)}\}_{L_j\in g_i})],
\]
where $\text{MaP}(\cdot)$ and $\text{MiP}(\cdot)$ are the element-wise max pooling and element-wise min pooling operators respectively.
\begin{figure*}[t]
\centering
\includegraphics[width=1.0\linewidth]{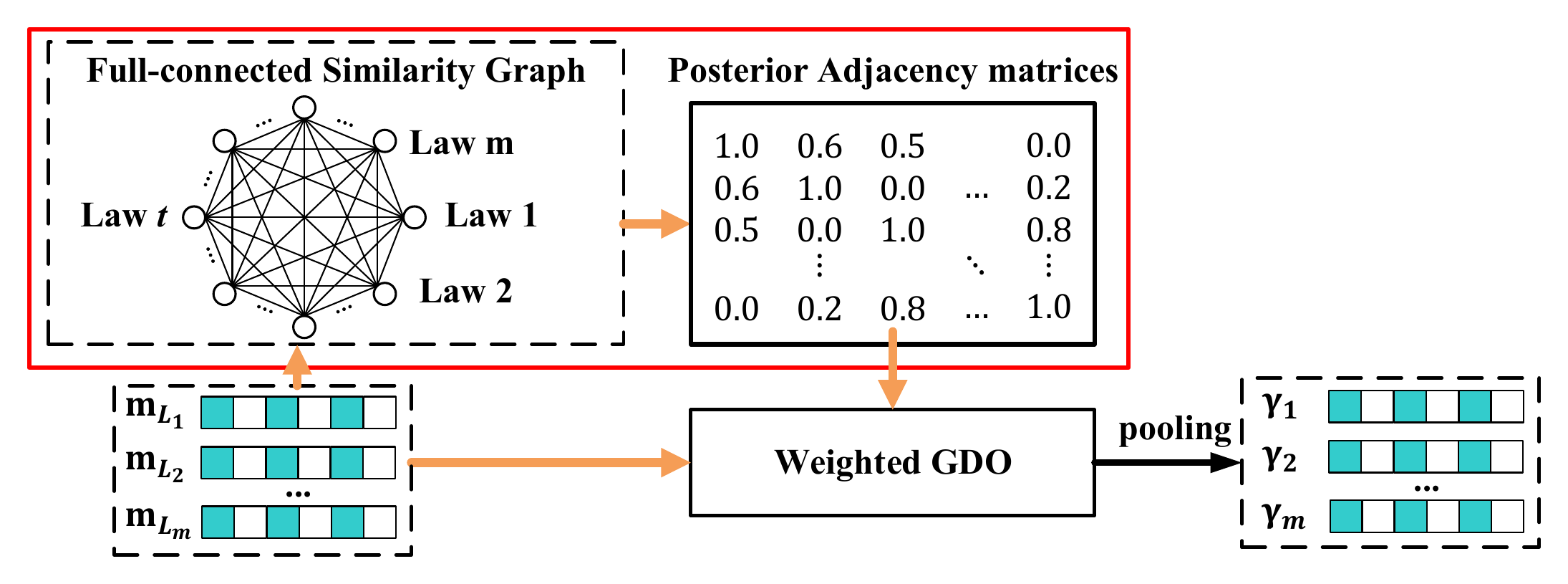}
\caption{Memory Distillation Module: this module constructs a weighted full-connected graph structure of law articles (or charges) and distills the distinguishable features of each law article (or charges) for attention calculation of the revised encoder.
}
\label{fig:Posterior_GDO}
\end{figure*}
\subsection{Distilling Revised Memories} \label{sec:memory_interaction}
The imbalance problem may cause the model's biased understanding of the similarity relation between law articles. To solve this issue, we design the \emph{revised memory mechanism} to dynamically sense the semantic similarity relationship between law articles that the model learned. Then, we propose a memory distillation module to learn the revised distinguishable representation (hereinafter called
\emph{revised distinction vector}) for each memory, relying on the \emph{fully-connected similarity graph} and the \emph{weighted graph distillation layer}.

\subsubsection{Revised Memory} To sense the semantic similarity relation of law articles that the model learned, we define a memory mechanism where each ``memory'' is associated with the corresponding law article, which is denoted as $\mathcal{M} = \{\mathbf{m}_{L_1}, \cdots, \mathbf{m}_{L_m}\}$, where $\mathbf{m}_{L_i} \in \mathbb{R}^{d_s}$ and $\mathcal{M} \in \mathbb{R}^{m \times d_s}$.

\subsubsection{Fully-connected Similarity Graph}
As memory is proposed to evaluate the posterior similarity between law articles, the similarity metric needs to be consistent with the metric function of the corresponding classifier. Here we show the cosine distance-based formula  (refer to Eq.~\ref{eq:metric_classifier}):
\begin{equation}\label{eq:metric_memory}
 a_{L_i,L_j} = \cos{(\mathbf{m}_{L_i}, \mathbf{m}_{L_j})} = 
 \frac{\mathbf{m}_{L_i}^\intercal \mathbf{m}_{L_j}}
 {\left \| \mathbf{m}_{L_i} \right \| \cdot \left \| \mathbf{m}_{L_j} \right \|}.
\end{equation}
After computing the scores of each memory pair, we get a fully-connected graph denoted by $G_M = \{\mathcal{M}, \mathcal{A}_{M}\}$, where the nodes represent the revised memories $\mathbf{m}_{L_i} \in \mathcal{M}$ and edges have the weights $a_{L_i, L_j} \in \mathcal{A}_{M}$.
The basic encoder and prior encoder enable the model the basic ability to distinguish confusion, and the revised encoder is used to sense the model's lack of detail in distinguishing confusion and to correct it.
Therefore, we need to maintain all the details on the fully-connected graph of revised memory instead of using grouping and other methods to prune it (refer to Sec.~\ref{sec:graph_prior}).

\subsubsection{Weighted Graph Distillation Layer} 
As the edge weights of the fully-connected graph reflect the similarity between law articles that the model learned, we use the weighted version GDO to utilize the edge weights and compute the revised distinction vector for each memory, inspired by the graph attention network~\cite{velivckovic2018GAT}. The computation formula is
\[
\alpha_{L_i, L_j} = 
\frac{\exp(a_{L_i,L_j})}
{\sum_{L_j\in \mathcal{L}\setminus \{ L_i \}} \exp(a_{L_i,L_j})},
\]
\[
 \mathbf{m}_{L_i}^{(l+1)} = \Phi_{\text{M}}^{(l)}\mathbf{m}_{L_i}^{(l)} - 
 \sum_{L_j\in \mathcal{L}\setminus \{ L_i \}} 
 \alpha_{L_i, L_j} \Psi_{\text{M}}^{(l)}(\mathbf{m}_{L_i}^{(l)} \oplus \mathbf{m}_{L_j}^{(l)}) + \mathbf{b}_{\text{M}}^{(l)},
\]
where $\alpha_{L_i, L_j}$ is the normalized weight and $\Phi_\text{M}^{(l)} \in \mathbb{R}^{d_{l+1} \times d_{l}},\Psi_\text{M}^{(l)} \in \mathbb{R}^{d_{l+1} \times 2d_{l}}$ are the trainable self-weighted matrix and the neighbor similarity extracting matrix for the memory distillation module, respectively. As the revised distinction vectors require a finer granularity to revise the biased understanding of the similarity between law articles caused by $ \mathbf{v}_{f}^{\text{b}} $ and $ \mathbf{v}_{f}^{\text{p}} $, we discard the setting of community and directly use the memory representation after $H$ layers as the final revised distinction vectors, i.e., $\mathbf{\gamma}_{i} = \mathbf{m}_{L_i}^{(H)}$.
Referring to the setting of existing memory networks~\cite{memory2016ICML, memory2018EMNLP}, we set the key vector $\mathbf{k}_{i}$ of each revised distinction vector $\gamma_{i}$ for querying. 
As the revised memories themselves have strong semantic characteristics, in D-LADAN, we simply set them as the key vectors, i.e., $ \mathbf{k}_i = \mathbf{m}_{L_i}^{(0)} = \mathbf{m}_{L_i} $.




Besides, it is worth noting that the revised memory mechanism is only related to the similarity relation between labels (i.e., law articles) that the model learned. Therefore, it is suitable for extending this mechanism to other sub-tasks (e.g., charge prediction and term of penalty prediction).

\subsection{Re-encoding Fact with Distinguishable Attention}
\label{sec:re_encoder}
In D-LADAN, the distinction vectors $\{\beta_i\}_{i=1}^k$ and $\{\gamma_{i}\}_{i=1}^{m}$ are computed to capture the corresponding distinguishable features from fact descriptions. Specifically, we first generate the most relevant distinction context vectors, $\hat{\beta}$ and $\hat{\gamma}$, based on the semantic correlation between the input fact description and the distinction vectors. Then, two similar re-encoders are used to generate the distinguishable representations of the input fact description relying on the $\hat{\beta}$ and $\hat{\gamma}$, respectively.

\subsubsection{Distinguishable Context Generation}
To capture a law case's prior distinguishable features from its fact description $f$, we define the following nonlinear function to compute the semantic correlation between it and all communities $g_{i}$ in the graph $G$:
\begin{equation}\label{eq:graph_selection}
\hat{\mathbf{X}} = \text{softmax}(\mathbf{W}_g\mathbf{v}_f^{\text{b}} + \mathbf{b}_g),
\end{equation}
where $\mathbf{v}_f^{\text{b}}$ is the basic representation of fact description $f$,
$\mathbf{W}_g \in \mathbb{R}^{k \times d_s}$ and $\mathbf{b}_g \in \mathbb{R}^k$ are the trainable weight matrix and bias respectively. Each element $\hat{X}_i \in \hat{\mathbf{X}}$, $i=1,...,k$ reflects the closeness between fact description $f$ and law articles community $g_i$.
The most relevant distinction vector $\hat{\beta}$ is computed as the weighted sum of all prior distinction vectors, that is:

\[\hat{\beta} =\sum_{i=1}^{k} \hat{X}_i \beta_i.\]

As for the revised distinguishable features, we use the following metric-based matching function to compute the semantic relativity between the input fact description and revised memories: 
\begin{equation}\label{eq:memory_matching}
S^{'}_i = \cos(
\mathbf{W}_k(\mathbf{v}_f^{\text{b}} \oplus \mathbf{v}_f^{\text{p}}), \mathbf{k}_i),
\end{equation}
where $\mathbf{W}_g \in \mathbb{R}^{d_s \times 2d_s}$ is the trainable weight matrix that maps fact representations into the same vector space as key vectors.
The most relevant revised distinction vector $\hat{\gamma}$ is computed by the same softmax function and weighted sum function with that prior distinction vectors $\hat{\beta}$ used, i.e.,
$ \hat{\mathbf{S}} = \text{softmax}([S^{'}_1, ..., S^{'}_m]) $ and $\hat{\gamma} =\sum_{i=1,\ldots, m} \hat{S}_i \gamma_i$ .
Then, the generated distinction vectors $\hat{\beta}$ and $\hat{\gamma}$ are input into the subsequent re-encoder for attentively extracting distinguishable features from fact description $f$.

\subsubsection{Fact Re-encoder}
Inspired by~\cite{yang2016hierarchical}, we attentively extract distinguishable features based on word-level and sentence-level Bi-directional Gated Recurrent Units (Bi-GRUs).
Since the calculation process is completely consistent, we only show the generation process of the prior representation $\mathbf{v}_f^{\text{p}}$.
Specifically, for each input sentence $S_i = [w_{i,1}, \cdots, w_{i, n_{i}}]$ in the fact description $f$,  word-level Bi-GRUs will output a hidden state sequence, that is,
\[ \mathbf{h}_{i,j} = [\overrightarrow{\text{GRU}}(\mathbf{w}_{i,j}), \overleftarrow{\text{GRU}}(\mathbf{w}_{i,j})],\ \ j = 1,..., n_i,\]
where $\mathbf{w}_{i,j}$ represents the word embedding of word $w_{i.j}$ and
$\mathbf{h}_{i,j} \in \mathbb{R}^{d_w}$.
Based on this hidden state sequence and the prior distinction vector $\hat{\beta}$, we calculate an attentive vector $[\alpha_{i,1},\ldots,\alpha_{i,n_i}]$,
where each $\alpha_{i,j}$ evaluates the discrimination ability of the corresponding word $w_{i,j} \in S_i$. $\alpha_{i,j}$ is formally computed as:
\[\alpha_{i,j} =\frac {\exp(\text{tanh}(\mathbf{W}_w{\mathbf{h}}_{i,j})^\mathsf{T}(\mathbf{W}_{gw}\hat{\beta}))} {\sum_{j}\exp(\text{tanh}(\mathbf{W}_w{\mathbf{h}}_{i,j}) ^\mathsf{T}(\mathbf{W}_{gw}\hat{\beta}))},\]
where $\mathbf{W}_{w}$ and $\mathbf{W}_{gw}$ are trainable weight matrices.
Then, we get a representation of sentence $S_i$ as:
\[\mathbf{v}_{s_i} = \sum_{j = 1}^{n_i}\alpha_{i,j}\mathbf{h}_{i,j},\]
where $n_i$ denotes the word number in sentence $S_i$.

By the above word-level Bi-GRUs, we get a sentence representations sequence $[\mathbf{v}_{s_1}, \ldots, \mathbf{v}_{s_{n_f}}]$,
where $n_f$ refers to the number of sentences in the fact description $f$.
Based on this sequence, similarly, we build sentence-level Bi-GRUs and calculate a sentence-level attentive vector $[\alpha_{1},\ldots,\alpha_{n_f}]$ that reflects the discrimination ability of each sentence, and then get the fact's prior representation $\mathbf{v}_f^{\text{p}} \in \mathbb{R}^{d_s}$.
Our sentence-level Bi-GRU is formulated as:
\[\mathbf{h}_{i} = [\overrightarrow{\text{GRU}}(\mathbf{v}_{s_i}),  \overleftarrow{\text{GRU}}(\mathbf{v}_{s_i})],\ \ i = 1,2,..., n_f, \]
\[\alpha_{i} =\frac {\exp(\text{tanh}(\mathbf{W_s}{\mathbf{h}}_{i})^\mathsf{T}(\mathbf{W}_{gs}\hat{\beta}))} {\sum_{i}\exp(\text{tanh}(\mathbf{W_s}{\mathbf{h}}_{i})^\mathsf{T}(\mathbf{W}_{gs}\hat{\beta}))},\]
\[
\mathbf{v}_f^{\text{p}} = \sum_{i=1}^{n_f}\alpha_{i}\mathbf{h}_{i}.
\]

As for the generation of $\mathbf{v}_f^{\text{r}}$, we replace the $\hat{\beta}$ in the above formulas with $\hat{\gamma}$ and use another set of trainable parameters.
Finally, we concatenate the basic representation $\mathbf{v}_f^{\text{b}}$, the prior representation $\mathbf{v}_f^{\text{p}}$ and the revised representation $\mathbf{v}_f^{\text{r}}$ as the final representation of fact description $f$, i.e., $\tilde{\mathbf{v}}_f = [\mathbf{v}_f^{\text{b}} \oplus \mathbf{v}_f^{\text{p}} \oplus \mathbf{v}_f^{\text{r}}]$.

\subsection{Prediction}\label{sec:prediction}

Based on $\tilde{\mathbf{v}}_f$,
we use the multi-task decoder to generate a corresponding feature vector $\tilde{\mathbf{v}}_f^i$ for each sub-task $ i \in \{l, c, t\}$, as mentioned in Sec.~\ref{sec:problem}, i.e., $l$: \textbf{law article prediction}; $c$: \textbf{charge prediction}; $t$: \textbf{term of penalty prediction}.
To obtain the prediction for each sub-task, we choose the metric-based classifier. Here we show the formula based on cosine distance consistent with the fully-connected graph computation in Sec.~\ref{sec:memory_interaction}:
\begin{equation}\label{eq:metric_classifier}
\hat{y}_{i} = \text{softmax}\left(\tau_i \cdot 
\frac{(\tilde{\mathbf{v}}_f^i)^\intercal \mathbf{W}_p^i}
{\left \| \tilde{\mathbf{v}}_f^i \right \| 
\cdot 
\left \| \mathbf{W}_p^i \right \|}\right), \ \  i \in \{l, c, t\}
\end{equation}
where $\mathbf{W}_p^i \in \mathbb{R}^{d_s \times \lvert Y_i\lvert}$ and $\tau_i$ are parameters specific to the corresponding sub-task and note the $\tau_i$ is a trainable scalar value.

\subsection{Training}\label{sec:training}
\noindent {\bf Loss Function.}
For training, we compute the cross-entropy loss function for each sub-task and take the loss sum of all sub-tasks as the overall prediction loss:
$$ \mathscr{L}_p = -\sum_{i} \sum_{j=1}^{\lvert Y_i\lvert} {y}_{i,j} \log(\hat{y}_{i,j}), \ \ i \in \{l, c, t\},$$
where $\lvert Y_i\lvert $ denotes the number of different classes (or, labels) for the corresponding sub-task and $[y_{i,1}, y_{i,2}, \dots, y_{i,\lvert Y_i\lvert}]$  refers to the one-hot ground-truth labels vector.
Besides, we also consider the loss of law article community selection (i.e., Eq.~(\ref{eq:graph_selection})) and the revised memory selection (i.e., Eq.~(\ref{eq:memory_matching})):
\[ 
\mathscr{L}_{c} = -\sum_{i=1}^{k} {X}_{i}\log({\hat{X}_{i}}); \ \
\mathscr{L}_{m} = -\sum_{i=1}^{m} {S}_{i}\log({\hat{S}_{i}}),\]
where $[X_1, X_2,\dots, X_k]$ and $[S_1, S_2,\dots, S_m]$ are separately the one-hot ground-truth vectors of the community and the revised memory, 
where only the element that covers the correctly applicable law article of input legal case is set to $1$ and others are $0$.
In summary, our final overall loss function is as follows:
\begin{equation}\label{eq:loss}
 \mathscr{Loss} = \mathscr{L}_p + \lambda_{c}\mathscr{L}_{c} + \lambda_{m}\mathscr{L}_{m},
\end{equation}
where $\lambda_{c}$ and $\lambda_{m}$ are the weight hyper-parameters.

\noindent {\bf Momentum Updating.}
As each row parameter of the metric-based classifier can be approximated as a prototype of the corresponding category~\cite{yang2018ConvolutionalPrototypeLearning}, we use the parameters of the law article classifier to update the revised memories with a momentum term $\lambda$ after each training step. For a given training step $t$, the update formula is
\[
\mathcal{M}^{(t)} = 
\lambda \mathcal{M}^{(t-1)} + (1 - \lambda){\mathbf{W}_p^{l}}^{(t)},
\]
where $\mathbf{W}_p^l$ denotes the parameters specific to the sub-task of law article prediction.
Notice that the classifiers and the revised memories need to choose a consistent metric for the momentum updating (cf., Eqs.~(\ref{eq:metric_memory}) and (\ref{eq:metric_classifier})).
Besides, as the revised memory mechanism is independent of prior knowledge, it can be configured for each sub-task. In this work, we also configured revised memory $\mathcal{M}_{c}$ for the charge prediction sub-task.

\noindent {\bf Implementation details.} Since the revised memory needs to roughly represent the similarity relationship between categories learned by the model at least, in the actual training, D-LADAN initializes the revised memory with the parameters of the corresponding classifier after a warm-up training. 
Assuming that warm-up training steps are $t_w$, then the revised memory is initialized by,
\[
\mathcal{M}^{(0)} = {\mathbf{W}_p^{l}}^{(t_w)}.
\]
Note that we freeze the revised representation $\mathbf{v}_f^{\text{r}}$ and only use the partial
loss (i.e., $\mathscr{Loss} = \mathscr{L}_p + 
\lambda_{c}\mathscr{L}_{c}$) for the warm-up training.

\subsection{The Upgraded Version: D-LADAN meets Transformers}\label{sec:Upgraded_Version}
From pre-trained models (PLMs)~\cite{cui2021pre,xiao2021lawformer,liang2024merge} to large language models (LLMs) ~\cite{brown2020gpt3,achiam2023gpt4}, Transformer-based architectures have enabled significant advances in the field of NLP and demonstrated their effectiveness in capturing context.
To further demonstrate the effectiveness of our proposed method, we propose an upgraded version of D-LADAN based on transformer-based architecture.
In this section, we use the BERT as an example to show how D-LADAN can be improved, which is denoted as $\text{D-LADAN}_{BERT}$.

Since PLMs have great advantages over the RNN models in long-distance text perception and context modeling, $\text{D-LADAN}_{BERT}$ discards the hierarchical design and models the case from the token level directly. 
Thus, fact description is treated as a sequence of tokens, i.e., $f = [t_1, \cdots, t_{n_s}]$, where $n_s$ is the sequence length. 
Taking the fact description as an input, $\text{D-LADAN}_{BERT}$ first uses the BERT model to get the hidden representation of each token, i.e.,
\[
[\mathbf{t}_1, \cdots, \mathbf{t}_{n_s}] = \text{BERT}([t_1, \cdots, t_{n_s}]),
\]
where $\mathbf{t}_i \in \mathbb{R}^{d_\textbf{BERT}}$ denotes the token representation of the token $t_{i}$ and $d_\textbf{BERT}$ is the output dimension of BERT.

Then, due to the proven ability of transformer layers to better capture long-range dependencies in language, we replace RNN layers with transformer layers in $\text{D-LADAN}_{BERT}$ to model the contextual information. 
The specific calculation formula is as follows, 
\[
[\mathbf{h}_1, \cdots, \mathbf{h}_{n_s}] = \text{Transformer}([\mathbf{t}_1, \cdots, \mathbf{t}_{n_s}]),
\]
where $\text{Transformer}(\cdot)$ denotes a transformer layer which consists of a 12-head self-attention layer and a feedforward layer and $\mathbf{h}_i \in \mathbb{R}^{d_\textbf{BERT}}$ is the hidden representation of the token $t_{i}$.

To unify the basic structure of the model, $\text{D-LADAN}_{BERT}$ uses a self-attention-like context attention layer to aggregate the token hidden representations to the fact representation.
We denote the context vector by $\mathbf{c}_{*}$, then the specific formula for the context attention layer is as follows,
\[
\mathbf{q}_{*} = \mathbf{W}_q\mathbf{c}_{*}; \ \ \mathbf{k}_i = \mathbf{W}_k\mathbf{h}_{i}; \ \ \mathbf{v}_i = \mathbf{W}_v\mathbf{h}_{i},
\]
\[
\alpha_{*,i} = \frac {\exp(\mathbf{q}_{*}^\mathsf{T}\mathbf{k}_i)} {\sum_{i}\exp(\mathbf{q}_{*}^\mathsf{T}\mathbf{k}_i)},
\]
\[
\mathbf{v}_f^{*} = \sum_{i=1}^{n_s}\alpha_{*,i}\mathbf{v}_{i},
\]
where $\mathbf{W}_q$, $\mathbf{W}_k$ and $\mathbf{W}_v$ are trainable weight matrices.
Following the design of D-LADAN, the corresponding context vector $\mathbf{c}_{b}$ is a trainable vector when computing the base representation $\mathbf{v}_f^{b}$.
As for the prior representation $\mathbf{v}_f^{p}$ and revised representation $\mathbf{v}_f^{r}$, the corresponding context vector are still the distinction vectors, i.e., $\mathbf{c}_{p}=\hat{\beta}$  and  $\mathbf{c}_{r}=\hat{\gamma}$.
\section{Experiments} \label{sec:experiment}

In this section, we introduce the setting of experiments and report the results with specific analyses.

\subsection{Datasets}

To verify the effectiveness of our method, we conduct experiments on two typical datasets: 1) the Chinese AI and Law challenge (CAIL2018) dataset~\cite{xiao2018cail2018} and 2) the Criminal dataset~\cite{hu2018few}. 
The statistics of these two datasets are shown in Table~\ref{tab:data}, and the detailed introduction is as follows:
\begin{itemize}
\item \textbf{CAIL2018\footnote{\url{http://cail.cipsc.org.cn/index.html}}~\cite{xiao2018cail2018}:} to evaluate the performance of our method, we use the two publicly available sub-datasets of the CAIL2018 dataset: \emph{CAIL-small} (the exercise stage dataset) and \emph{CAIL-big} (the first stage dataset).
The case samples in both datasets contain fact descriptions, applicable law articles, charges, and the term of penalty.
As for data processing, we first filter out samples with fewer than ten meaningful words.
To be consistent with state-of-the-art methods, we filter out the case samples with multiple applicable law articles and multiple charges.
Meanwhile, referring to~\cite{zhong2018legal}, we only keep the law article and the charge that applies to not less than 100 corresponding case samples and divide the terms of penalty into 11 non-overlapping intervals.

\item \textbf{Criminal\footnote{\url{https://github.com/thunlp/attribute_charge.}}~\cite{hu2018few}:}
to further prove the ability of D-LADAN to solve the imbalance problem as well as the confusing law article (or charge) problem, we evaluate it on the available datasets from ~\cite{hu2018few}, which contains real cases for few-shot charges prediction. The dataset has three subsets of different sizes, denoted as Criminal-S (small), Criminal-M (medium), and Criminal-L (large).
These datasets also filtered cases involving multiple defendants and multiple charges. 
\end{itemize}

\begin{table}[t]
\caption{
Statistics of all experimental datasets. "--" denotes the Criminal datasets have no labels about law articles and the term of penalty.
}\label{tab:data}
\centering
\begin{tabular}{lrrrrr}
\toprule
Dataset & CAIL-small & CAIL-big & Criminal-S    & Criminal-M    & Criminal-L\\
\midrule
\#Training Set Cases  & 101,619    & 1,587,979  & 61,586    & 153,518   & 306,890 \\
\#Test Set Cases  & 26,749    & 185,120  & 7,702    & 19,188   & 38,366\\
\#Law Articles  & 103   & 118  & --   & --   & --\\
\#Charges     & 119     & 130  & 149   & 149   & 149\\
\#Term of Penalty     & 11    & 11  & --   & --   & --\\
\bottomrule
\end{tabular}
\end{table}

 \subsection{Baselines and Settings}
\noindent   {\bf Baselines.} We compare D-LADAN with baselines including:
\begin{itemize}

\item \textbf{CNN~\cite{kim2014convolutional}:} a CNN-based model with multiple filter window widths for text classification. 

\item \textbf{HARNN~\cite{yang2016hierarchical}:} an RNN-based neural network with a hierarchical attention mechanism for document classification.

\item \textbf{FLA~\cite{luo2017learning}:} a charge prediction method considering the interaction between fact description and applicable laws. 
It points out that law articles can help filter out key information related to jurisprudence in a legal case as the judges must cite the applicable law articles to determine the final charges. Thus, FLA first proposes a retrieval method to select the top-$k$ relevant law articles for each case. Then, FLA uses the attention mechanism to capture the legal-related information of legal cases based on the context vectors of selected law articles. Finally, the extracted vector representation is employed to predict the charges.

\item \textbf{Few-Shot~\cite{hu2018few}:} a deep neural network-based model aims to solve both problems of the few-shot charges and the confusing charges.
It manually annotates ten distinguishable attributes of charges to enhance the relation between fact description and charges.
In practice, it constructs the attribute prediction sub-task and proposes the attribute-aware attention mechanism to enhance the charge-related semantic information of fact descriptions.

\item \textbf{TOPJUDGE~\cite{zhong2018legal}:} a topological multi-task learning framework that considers the relationship between three sub-tasks in LJP. This model formalizes the explicit dependencies over sub-tasks in a directed acyclic graph, following the judge's judgment norms under the statutory system, in which the judge first evaluates the possible violation of the law, then determines the crime, and finally judges the punishment according to the law.

\item \textbf{MPBFN-WCA~\cite{yang2019legal}:} another multi-task learning framework for LJP task. Compared with the TOPJUDGE~\cite{zhong2018legal}, its innovation lies in the backward verification framework besides the typical forward dependency, which is inspired by the assumption that after making a decision, judges should analyze and confirm whether the charges and penalty conditions are following the provisions of the law articles.

\item \textbf{LADAN~\cite{xu2020Ladan}:} 
the proposed method of our conference version, which only considers the prior relationships between law articles and uses the GDO to determine the differences between similar law articles. In other words, such a model only combines the basic representation and the prior representation as the final representation of the fact description, i.e., $\tilde{\mathbf{v}}_f = [\mathbf{v}_f^{\text{b}} \oplus \mathbf{v}_f^{\text{p}}]$.

\item \textbf{GFDN~\cite{zhao2022GFDN}:} a modified method of the LADAN model which transforms the fact description to a graph structure from a sequence, and then uses the graph convolutional network as the basic encoder to get a better representation.

\item \textbf{R-former
\footnote{\url{https://github.com/DQ0408/R-former}}~\cite{dong2021R_former}:} a relational learning-based method, which considers the consistency between the labels that belong to different tasks. This method maps all label-aware representations of the input fact into a joint space and treats LJP as a node classification task to output the prediction result. Notice that this method used the large-scale pre-trained model (vanilla Transformer~\cite{vaswani2017Transformer}) for encoding fact description. To ensure fairness, we reimplement it by replacing the Transformer with HARNN as the encoder in all experiments.

\item \textbf{Neurjudge
\footnote{\url{https://github.com/bigdata-ustc/NeurJudge}}~\cite{yue2021Neurjudge}:} a circumstance-aware neural framework for JLP task, which considers functional differences in different parts of a legal case's fact description. 
It leverages the intermediate results of sub-tasks to decouple the fact description into several separate circumstances and exploits them to predict the results of other sub-tasks, which is inspired by judges' behavior that uses different parts of the fact descriptions of a legal case to decide different trial items.

\item \textbf{CEEN}
\footnote{\url{https://github.com/lvyougang/CEEN/}}~\cite{lyu2022CEEN}: a reinforcement learning-based method, which considers enhancing the fact representation with legal concepts in the JLP task.
It constructs four types of criminal elements and tags manual labels for each training sample.
With a reinforcement learning-based element discriminator, it takes the fact representation to perceive legal element information, thereby improving performance on LJP tasks.

\item \textbf{CL4LJP~\cite{zhang2023CL4LJP}:} a supervised contrastive learning framework to solve the confusing law articles and charges.
It extends prior knowledge of law articles and charges to the instance level by selecting negative samples in contrast to learning based on the relationships of law articles and charges, including taking legal cases belonging to different items in the same chapter of the Criminal Law as negative samples and legal cases belonging to the same relevant law articles but different charges as negative samples.

\item \textbf{BERT~\cite{cui2021pre}}: a Transformer-based method that is pre-trained on Chinese Wikipedia documents.

\end{itemize}

\begin{table*}[t]
\centering
\caption{
Judgment prediction results on CAIL-small. The results of our D-LADAN are in \textbf{bold}, and the best results of baselines are \underline{underlined}. The best results are \underline{\textbf{bold with underline}}. $\dagger$ denotes D-LADAN achieves significant improvements over all existing baselines in paired t-test with $p$-value $< 0.05$.
}\label{tab:CAIL-small}
\resizebox{\textwidth}{!}{%
\begin{tabular}{lcccccccccccc}
\toprule
 Tasks              & \multicolumn{4}{c}{Law Articles}                     & \multicolumn{4}{c}{Charges}                          & \multicolumn{4}{c}{Term of Penalty}                   \\ \cmidrule(lr){2-5} \cmidrule(lr){6-9} \cmidrule(lr){10-13}
 Metrics            & Acc.        & MP          & MR          & F1
                    & Acc.        & MP          & MR          & F1
                    & Acc.        & MP          & MR          & F1          \\
 \midrule
\multicolumn{13}{c}{\textbf{Simple Backbone}} \\
 \midrule
 FLA+MTL            & $77.74$     & $75.32$     & $74.36$     & $72.93$
                    & $80.90$     & $79.25$     & $77.61$     & $76.94$
                    & $36.48$     & $30.94$     & $28.40$     & $28.00$     \\

 CNN+MTL            & $78.71$     & $76.02$     & $74.87$     & $73.79$
                    & $82.41$     & $81.51$     & $79.34$     & $79.61$
                    & $35.40$     & $33.07$     & $29.26$     & $29.86$     \\

 HARNN+MTL          & $79.79$     & $75.26$     & $76.79$     & $74.90$
                    & $83.80$     & $82.44$     & $82.78$     & $82.12$
                    & $36.17$     & $34.66$     & $31.26$     & $31.40$     \\

 Few-Shot+MTL       & $79.30$     & $77.80$     & $77.59$     & $76.09$
                    & $83.65$     & $80.84$     & $82.01$     & $81.55$
                    & $36.52$     & $35.07$     & $26.88$     & $27.14$     \\

 TOPJUDGE (TOP)     & $79.88$     & $79.77$     & $73.67$     & $73.60$
                    & $82.10$     & $83.60$     & $78.42$     & $79.05$
                    & $36.29$     & $34.73$     & $32.73$     & $29.43$     \\

 MPBFN-WCA          & $79.12$     & $76.30$     & $76.02$     & $74.78$
                    & $82.14$     & $82.28$     & $80.72$     & $80.72$
                    & $36.02$     & $31.94$     & $28.60$     & $29.85$     \\

 GFDN               & $81.37$     & $78.07$     & $77.43$     & $76.67$
                    & $84.83$     & $83.96$     & $82.73$     & $82.93$
                    & $38.38$     & $36.46$     & $32.88$     & $32.97$     \\
 
 $\text{R-former}_{simple}$  & $ 81.92 $     & $ 78.54 $     & $ 78.48 $     & $ 77.16 $
                    & $ \underline{85.83} $     & $ 84.56 $     & $ 83.25 $     & $ 83.34 $
                    & $ 38.88 $     & $ 36.63 $     & $ 33.54 $     & $ 34.43 $     \\
                     
  Neurjudge          & $ \underline{82.30} $     & $ \underline{78.89} $     & $ \underline{79.35} $     & $ \underline{78.28} $
                      & $ 85.60 $     & $ \underline{84.80} $     & $ \underline{84.15} $     & $ \underline{84.10} $
                      & $ \underline{38.90} $     & $ \underline{36.84} $     & $ \underline{33.96} $     & $ \underline{34.58} $     \\

  CL4JLP             & $ 81.63 $     & $ 77.21 $     & $ 77.97 $     & $ 76.48 $
                     & $ 84.48 $     & $ 83.38 $     & $ 84.06 $     & $ 83.19 $
                     & $ 37.77 $     & $ 36.23 $     & $ 31.76 $     & $ 31.81 $     \\
 \midrule
 LADAN+MTL          & $81.20$     & $78.24$     & $77.38$     & $76.47$
                    & $85.07$     & $83.42$     & $82.52$     & $82.74$
                    & $38.29$     & $36.16$     & $32.49$     & $32.65$     \\

\bf{D-LADAN+MTL}    & $ \underline{\bf 83.24}^{\dagger} $ & $ \underline{\bf 80.32}^{\dagger} $ 
                    & $ \underline{\bf 81.46}^{\dagger} $ & $ \underline{\bf 79.73}^{\dagger} $
                    & $ \bf 86.95^{\dagger} $ & $ \bf 85.69^{\dagger} $ 
                    & $ \bf 86.17^{\dagger} $ & $ \bf 85.48^{\dagger} $
                    & $ \underline{\bf 40.97}^{\dagger} $ & $ \bf 38.31^{\dagger} $ 
                    & $ \bf 37.06^{\dagger} $ & $ \bf 36.89^{\dagger} $ \\

\bf{D-LADAN+TOP}   & $ \bf 83.05^{\dagger} $ & $ \bf 79.22 $ 
                        & $ \bf 80.71^{\dagger} $ & $ \bf 78.79^{\dagger} $
                        & $ \underline{\bf 87.29}^{\dagger} $ & $ \underline{\bf 85.70}^{\dagger} $ & $ \underline{\bf 86.38}^{\dagger} $ & $ \underline{\bf 85.71}^{\dagger} $
                        & $ \bf 40.88^{\dagger} $ & $ \underline{\bf 38.35}^{\dagger} $ 
                        & $ \underline{\bf 37.22}^{\dagger} $ & $ \underline{\bf 37.36}^{\dagger} $ \\

\bf{D-LADAN+MPBFN}      & $ \bf 82.77^{\dagger} $ & $ \bf 79.32 $ 
                        & $ \bf 80.60^{\dagger} $ & $ \bf 78.76^{\dagger} $
                        & $ \bf 87.01^{\dagger} $ & $ \bf 85.25 $ 
                        & $ \bf 86.34^{\dagger} $ & $ \bf 85.46^{\dagger} $
                        & $ \bf 40.68^{\dagger} $ & $ \bf 37.52^{\dagger} $ 
                        & $ \bf 36.32^{\dagger} $ & $ \bf 36.36^{\dagger} $ \\
 \midrule 
\multicolumn{13}{c}{\textbf{Transformer Backbone}} \\
\midrule
BERT         & $ 83.73 $     & $ 80.87 $     & $ 81.75 $     & $ 80.10 $
                    & $ 87.48 $     & $ 86.21 $     & $ 85.75 $     & $ 85.69 $
                    & $ 41.16 $     & $ 39.91 $     & $ 38.08 $     & $ 38.37 $     \\
R-former           
                    & $ 84.49 $     & $ 82.17 $     
                    & $ 82.08 $     & $ 81.16 $
                    & $ \underline{89.09} $     & $ \underline{88.41} $    
                    & $ 88.00 $     & $ 87.80 $
                    & $ 42.19 $     & $ 40.56 $     
                    & $ \underline{39.56} $     & $ 39.47 $     \\
CEEN     
                    & $ \underline{84.56} $     & $ \underline{82.77} $     
                    & $ \underline{82.83} $     & $ \underline{81.51} $
                    & $ 88.77 $     & $ 88.19 $     
                    & $ \underline{88.38} $     & $ \underline{87.97} $
                    & $ \underline{42.54} $     & $ \underline{40.73} $     
                    & $ 39.51 $     & $ \underline{39.70} $     \\
\midrule
\bf $\text{D-LADAN}_{BERT}$    
                    & $ \underline{\bf 85.88}^{\dagger} $ & $ \underline{\bf 84.56}^{\dagger} $ 
                    & $ \underline{\bf 83.52}^{\dagger} $ & $ \underline{\bf 82.38}^{\dagger} $
                    & $ \underline{\bf 89.93}^{\dagger} $ & $ \underline{\bf 88.78}^{\dagger} $ 
                    & $ \underline{\bf 88.79}^{\dagger} $ & $ \underline{\bf 88.48}^{\dagger} $
                    & $ \underline{\bf 43.33}^{\dagger} $ & $ \underline{\bf 42.74}^{\dagger} $ 
                    & $ \underline{\bf 40.07}^{\dagger} $ & $ \underline{\bf 40.90}^{\dagger} $ \\
 \bottomrule
\end{tabular}%
}
\end{table*}

Similar to existing works~\cite{luo2017learning,zhong2018legal}, we train the baselines CNN, HARNN, FLA, GFDN, and LADAN using a multi-task framework (recorded as MTL) and select a set of the best experimental parameters according to the range of the parameters given in their original papers.
Besides, we use our method D-LADAN with the same multi-task framework (including MTL, TOPJUDGE, and MPBFN) to demonstrate our superiority in feature extraction.
When Criminal datasets only focus on the single charge prediction sub-task of about 149 charges, TOPJUDGE and MPBFN degenerate into CNN, R-former and CEEN degrade into BERT and Neurjudge degrades into Bi-GRU due to the lack of the relation between sub-tasks.
Due to the same reason, the CL4LJP would only keep the basic label-aware negative samples, where the negative legal cases belong to different items in the same chapter of the
Criminal Law as negative samples and the ones belong to the same relevant law articles
but different charges are not suitable for such a single task setting.
In contrast, our D-LADAN with the law article prior knowledge\footnote{The prior knowledge about law articles of the Criminal dataset can be obtained by reverse-indexing the relationship between articles of law and charges in the PRC Criminal Law} and the charge-related revised memory mechanism can still be suitable for solving the single charge prediction sub-task.

\begin{table*}[t]
\centering
\caption{
Judgment prediction results on CAIL-big. The results of our D-LADAN are in \textbf{bold}, and the best results of baselines are \underline{underlined}. The best results are \underline{\textbf{bold with underline}}. $\dagger$ denotes D-LADAN achieves significant improvements over all existing baselines in paired t-test with $p$-value $< 0.05$.
}\label{tab:CAIL-big}
\resizebox{\textwidth}{!}{%
\begin{tabular}{lcccccccccccc}
\toprule
 Tasks              & \multicolumn{4}{c}{Law Articles}                     & \multicolumn{4}{c}{Charges}                          & \multicolumn{4}{c}{Term of Penalty}                   \\ 
 \cmidrule(lr){2-5} \cmidrule(lr){6-9} \cmidrule(lr){10-13}
 Metrics            & Acc.        & MP          & MR          & F1
                    & Acc.        & MP          & MR          & F1
                    & Acc.        & MP          & MR          & F1          \\
 \midrule
\multicolumn{13}{c}{\textbf{Simple Backbone}} \\
 \midrule
 FLA+MTL            & $93.23$     & $72.78$     & $64.30$     & $66.56$
                    & $92.76$     & $76.35$     & $68.48$     & $70.74$
                    & $57.63$     & $48.93$     & $45.00$     & $46.54$     \\

 CNN+MTL            & $95.84$     & $83.20$     & $75.31$     & $77.47$
                    & $95.74$     & $86.49$     & $79.00$     & $81.37$
                    & $55.43$     & $45.13$     & $38.85$     & $39.89$     \\

 HARNN+MTL          & $95.63$     & $81.48$     & $74.57$     & $77.13$
                    & $95.58$     & $85.59$     & $79.55$     & $81.88$
                    & $57.38$     & $43.50$     & $40.79$     & $42.00$     \\

 Few-Shot+MTL       & $96.12$     & $85.43$     & $80.07$     & $81.49$
                    & $96.04$     & $88.30$     & $80.46$     & $83.88$
                    & $57.84$     & $47.27$     & $42.55$     & $43.44$     \\

 TOPJUDGE (TOP)     & $95.85$     & $84.84$     & $74.53$     & $77.50$
                    & $95.78$     & $86.46$     & $78.51$     & $81.33$
                    & $57.34$     & $47.32$     & $42.77$     & $44.05$     \\

 MPBFN-WCA          & $96.06$     & $85.25$     & $74.82$     & $78.36$
                    & $95.98$     & $89.16$     & $79.73$     & $83.20$
                    & $58.14$     & $45.86$     & $39.07$     & $41.39$     \\

 GFDN               & $96.60$     & $86.25$     & $80.80$     & $82.40$
                    & $96.55$     & $88.54$     & $83.91$     & $85.46$
                    & $59.68$     & $51.81$     & $45.36$     & $46.99$     \\
 
 $\text{R-former}_{simple}$  & $ 96.69 $     & $ 86.62 $     
                    & $ 81.38 $     & $ \underline{83.02} $
                    & $ \underline{96.65} $     & $ 89.01 $     
                    & $ \underline{84.43} $     & $ \underline{85.88} $
                    & $ 59.86 $     & $ \underline{52.23} $     
                    & $ 46.01 $     & $ 47.38 $     \\
                    
                     
  Neurjudge         & $ \underline{96.76} $     & $ 86.25 $     
                    & $ \underline{81.39} $     & $ 82.89 $
                    & $ 96.11 $     & $ \underline{89.56} $     
                    & $ 80.42 $     & $ 83.72 $
                    & $ \underline{60.19} $     & $ 51.76 $     
                    & $ \underline{46.31} $     & $ \underline{47.97} $     \\
                     
  CL4JLP            & $ 96.43 $     & $ \underline{86.96} $     
                    & $ 77.00 $     & $ 80.09 $
                    & $ 96.39 $     & $ 88.86 $   
                    & $ 81.62 $     & $ 84.33 $
                    & $ 57.50 $     & $ 49.55 $     
                    & $ 40.34 $     & $ 41.70 $     \\
                     
 \midrule
 LADAN+MTL          & $96.57$     & $86.22$     & $80.78$     & $82.36$
                    & $96.45$     & $88.51$     & $83.73$     & $85.35$
                    & $59.66$     & $51.78$     & $45.34$     & $46.93$     \\

 \bf D-LADAN+MTL        & $ \bf 96.95 $ & $ \underline{\bf 87.22} $ 
                        & $ \underline{\bf 83.17}^{\dagger} $ & $ \underline{\bf 84.66}^{\dagger} $
                        & $ \underline{\bf 96.90} $ & $ \underline{\bf 90.19} $ 
                        & $ \underline{\bf 86.09}^{\dagger} $ & $ \underline{\bf 87.71}^{\dagger} $
                        & $ \bf 60.92^{\dagger} $ & $ \underline{\bf 52.88} $ 
                        & $ \underline{\bf 48.57}^{\dagger} $ & $ \bf 49.94^{\dagger} $ \\

 \bf D-LADAN+TOP   & $ \underline{\bf 96.97} $ & $ \bf 87.01 $ 
                        & $ \bf 82.98^{\dagger} $ & $ \bf 84.35^{\dagger} $
                        & $ \bf 96.86 $ & $ \bf 89.83 $ 
                        & $ \bf 85.90^{\dagger} $ & $ \bf 87.31^{\dagger} $
                        & $ \underline{\bf 61.01}^{\dagger} $ & $ \bf 52.54 $ 
                        & $ \bf 48.52^{\dagger} $ & $ \underline{\bf 50.02}^{\dagger} $ \\
                        
 \bf D-LADAN+MPBFN      & $ \bf 96.93 $ & $ \bf 87.11 $ 
                        & $ \bf 82.79^{\dagger} $ & $ \bf 84.25^{\dagger} $
                        & $ \bf 96.80 $ & $ \bf 89.68 $ 
                        & $ \bf 85.65^{\dagger} $ & $ \bf 87.25^{\dagger} $
                        & $ \bf 60.60^{\dagger} $ & $ \bf 52.77 $ 
                        & $ \bf 48.44^{\dagger} $ & $ \bf 49.75^{\dagger} $ \\
\midrule
\multicolumn{13}{c}{\textbf{Transformer Backbone}} \\
\midrule
BERT         
                    & $ 97.10 $     & $ 87.35 $     & $ 84.12 $     & $ 84.42 $
                    & $ 97.11 $     & $ 91.12 $     & $ 88.28 $     & $ 89.36 $
                    & $ 60.41 $     & $ 51.34 $     & $ 49.92 $     & $ 51.57 $     \\
R-former           
                    & $ 97.54 $     & $ 88.42 $     
                    & $ 84.52 $     & $ 85.94 $
                    & $ \underline{97.56} $     & $ \underline{92.61} $     
                    & $ \underline{89.27} $     & $ \underline{90.64}$
                    & $ 60.74 $     & $ 52.93 $     
                    & $ \underline{53.12} $     & $ 52.74 $     \\

CEEN      
                    & $ \underline{97.43} $     & $ \underline{89.14} $     
                    & $ \underline{86.15} $     & $\underline{86.88}$
                    & $ 97.40 $     & $ 91.34 $     
                    & $ 89.04 $     & $ 89.86 $
                    & $ \underline{62.89} $     & $ \underline{54.85} $     
                    & $ 52.39 $     & $ \underline{53.20} $     \\
\midrule
\bf $\text{D-LADAN}_{BERT}$   
                & $ \underline{\bf 97.65} $ & $ \underline{\bf 89.51} $ 
                & $ \underline{\bf 86.75}^{\dagger} $ & $ \underline{\bf 87.64}^{\dagger} $
                & $ \underline{\bf 97.64} $ & $ \underline{\bf 92.53} $ 
                & $ \underline{\bf 90.11}^{\dagger} $ & $ \underline{\bf 90.95} $
                & $ \underline{\bf 64.72}^{\dagger} $ & $ \underline{\bf 56.77}^{\dagger} $ 
                & $ \underline{\bf 55.98}^{\dagger} $ & $ \underline{\bf 56.21}^{\dagger} $ \\
 \bottomrule
\end{tabular}%
}
\end{table*}
\noindent   {\bf Experimental Settings.} 
For models without a Transformer-based encoder, we use the THULAC~\cite{Thulac2016} tool to get the word segmentation because all case samples are in Chinese.
Afterward, we use the Skip-Gram model~\cite{mikolov2013distributed} to pre-train word embeddings on these case documents, where the model's embedding size and frequency threshold are set to 200 and 25 respectively\footnote{As for the Criminal datasets, the embedding vectors are available in the corresponding GitHub link.}.
For models with a Transformer-based encoder, we use the pre-trained weights and the hidden size
is 768. 
Meanwhile, we set the maximum document length as 512 for CNN-based models in baselines and set the maximum sentence length to 100 words and the maximum document length to 15 sentences for LSTM-based models.
For the Transformer-based model, following the set of R-formers, we set the maximum document length as 512 tokens.
For samples that exceed the text limit we select the concatenation of the first 255 tokens and the last 255 tokens as input
As for hyper-parameters setting, we set the dimension of all latent states (i.e., $d_w$, $d_s$, $d_l$, and $d_f$) as 256, the threshold $\tau$ as $0.35$, and the momentum term $\lambda$ as 0.9.
In our method D-LADAN, we use two graph distillation layers, and a Bi-GRU with a randomly initialized attention vector $u$ is adopted as the basic document encoder.
For training, we set the learning rate of the Adam optimizer to 0.001, the batch size to 128, $\lambda_{c}$ to $0.1$, and $\lambda_{m}$ to $0.1$.
After training every model for 32 epochs, we choose the best model on the validation set for testing.
For all methods, we performed 10 repeat experiments and averaged them for comparison.

\subsection{Basic Performance Evaluation}
To compare the performance of the baselines and our methods,
we choose four metrics widely used for multi-classification tasks, including accuracy (Acc.), macro-precision (MP), macro-recall (MR), and macro-F1 (F1).
Since the confusing law article (or charges) issue often occurs between a few categories and both CAIL and Criminal datasets are quite imbalanced, we mainly evaluate all methods with the F1 score, which more objectively reflects the effectiveness of our D-LADAN and other baselines.
Table~\ref{tab:CAIL-small}, ~\ref{tab:CAIL-big}, and ~\ref{tab:Criminal-all} shows the experimental results on CAIL-small, CAIL-big, and Criminal datasets, respectively. 
Our D-LADAN performs the best in terms of all evaluation metrics.
Compared with the state-of-the-art R-former model,
our D-LADAN improved the F1-scores of law article prediction, charge prediction, and the term of penalty prediction on dataset CAIL-small by $ 0.87 $\%, $0.51$\%, and $1.20$\% respectively, and about $0.76$\%, $0.31$\%, and $3.01$\% on dataset CAIL-big.
For the three Criminal datasets, D-LADAN improves the F1-score by $2.93$\%, $2.20$\%, and $0.82$\%, respectively.
We also get some confirmatory observations:
\begin{table*}[t]
\centering
\small
\caption{
Judgment prediction results on Criminal datasets. The results of our D-LADAN are in \textbf{bold}, and the best results of baselines are \underline{underlined}. $\dagger$ denotes D-LADAN achieves significant improvements over all existing baselines in paired t-test with $p$-value $< 0.05$.
}\label{tab:Criminal-all}
\resizebox{\textwidth}{!}{%
\begin{tabular}{lcccccccccccc}
\toprule
 Datasets              & \multicolumn{4}{c}{Criminal-S}                     & \multicolumn{4}{c}{Criminal-M}                          & \multicolumn{4}{c}{Criminal-L}                   \\ 
 \cmidrule(lr){2-5} \cmidrule(lr){6-9} \cmidrule(lr){10-13}
 Metrics            & Acc.        & MP          & MR          & F1
                    & Acc.        & MP          & MR          & F1
                    & Acc.        & MP          & MR          & F1          \\
 \midrule
\multicolumn{13}{c}{\textbf{Simple Backbone}} \\
 \midrule
 CNN            & $ 91.92 $     & $ 50.53 $     & $ 44.90 $     & $ 46.13 $
                & $ 93.53 $     & $ 57.61 $     & $ 48.11 $     & $ 50.53 $
                & $ 93.91 $     & $ 66.02 $     & $ 50.32 $     & $ 54.74 $     \\

 HARNN          & $ 92.69 $     & $ 60.01 $     & $ 58.38 $     & $ 56.95 $
                & $ 94.66 $     & $ 65.78 $     & $ 63.04 $     & $ 62.62 $
                & $ 95.08 $     & $ 72.83 $     & $ 66.66 $     & $ 67.91 $     \\
                
 FLA            & $ 92.82 $     & $ 57.04 $     & $ 53.91 $     & $ 53.41 $
                & $ 94.68 $     & $ 66.72 $     & $ 60.36 $     & $ 61.83 $
                & $ 95.71 $     & $ 73.27 $     & $ 67.10 $     & $ 68.58 $     \\

 Few-Shot       & $ 93.41 $     & $ 66.74 $     & $ 69.23 $     & $ 64.90 $
                & $ 94.39 $     & $ 68.31 $     & $ 69.19 $     & $ 67.14 $
                & $ 95.81 $     & $ 75.76 $     & $ 73.74 $     & $ 73.08 $     \\

 GFDN           & $ \underline{94.92} $     & $ \underline{71.26} $     & $ \underline{69.77} $     & $ \underline{69.58} $
                & $ 95.85 $     & $ \underline{76.46} $     & $ \underline{73.24} $     & $ \underline{73.55} $
                & $ \underline{96.45} $     & $ \underline{81.80} $     & $ \underline{76.75} $     & $ \underline{78.12} $     \\
                
 NeurJudge      & $ 94.26 $     & $ 65.61 $     & $ 62.68 $     & $ 63.27 $
                & $ 94.84 $     & $ 67.77 $     & $ 64.01 $     & $ 64.62 $
                & $ 95.70 $     & $ 78.85 $     & $ 72.26 $     & $ 74.10 $     \\

 CL4LJP         & $ 93.24 $     & $ 56.03 $     & $ 53.10 $     & $ 53.55 $
                & $ 94.86 $     & $ 65.99 $     & $ 63.34 $     & $ 63.22 $
                & $ 95.33 $     & $ 72.47 $     & $ 66.26 $     & $ 67.98 $     \\

 \midrule
 LADAN          & $ 94.88 $     & $ 71.15 $     & $ 69.26 $     & $ 69.04 $
                & $ \underline{95.87} $     & $ 75.93 $     & $ 72.75 $     & $ 73.14 $
                & $ 96.41 $     & $ 81.73 $     & $ 76.28 $     & $ 77.99 $     \\

 \bf D-LADAN    & $\bf 95.16^{\dagger} $ & $\bf 74.16^{\dagger} $ 
                & $\bf 72.75^{\dagger} $ & $\bf 72.51^{\dagger} $
                & $\bf 95.98 $ & $\bf 78.04^{\dagger} $ 
                & $\bf 75.95^{\dagger} $ & $\bf 75.75^{\dagger} $
                & $\bf 96.55 $ & $\bf 83.24^{\dagger} $ 
                & $\bf 77.45^{\dagger} $ & $\bf 78.94^{\dagger} $ \\

\midrule
\multicolumn{13}{c}{\textbf{Transformer Backbone}} \\
\midrule
BERT         
                    & $ \underline{95.78} $     & $ \underline{74.92} $     
                    & $ \underline{74.04} $     & $ \underline{73.95} $
                    & $ \underline{96.44} $     & $ \underline{80.97} $     
                    & $ \underline{79.00} $     & $ \underline{78.91} $
                    & $ \underline{97.03} $     & $ \underline{88.76} $     
                    & $ \underline{82.22} $     & $ \underline{84.22} $     \\

\midrule
\bf $\text{D-LADAN}_{BERT}$   
                    & $\bf 95.97 $ & $\bf 76.27^{\dagger} $ 
                    & $\bf 77.83^{\dagger} $ & $\bf 76.16^{\dagger} $
                    & $\bf 96.59 $ & $\bf 82.05^{\dagger} $ 
                    & $\bf 81.64^{\dagger} $ & $\bf 80.55^{\dagger} $
                    & $\bf 97.12 $ & $\bf 89.63^{\dagger} $ 
                    & $\bf 84.37^{\dagger} $ & $\bf 84.85^{\dagger} $ \\
\bottomrule
\end{tabular}%
}
\end{table*}
\begin{enumerate}
\item{The comparison under the same multi-task framework (i.e., MTL, TOPJUDGE, and MPBFN) shows that our D-LADAN extracted more effective features from fact descriptions than all baselines (cf. Tables~\ref{tab:CAIL-small} and~\ref{tab:CAIL-big}).}

\item{Compared with the previous LADAN, Tables~\ref{tab:CAIL-small},~\ref{tab:CAIL-big},  and~\ref{tab:Criminal-all} show that our D-LADAN tends to improve the MR score more than the MP score. This phenomenon potentially reflects that the revised memory mechanism is more inclined to improve the performance of the categories with few training samples. (see Sec.~\ref{subsec: Imbalance Expreiment} for detail)}

\item{Compared with the simple backbone, D-LADAN shows more improvement than baselines under the BERT backbone in the penalty prediction sub-task, which indicates that D-LADAN benefits more from BERT's improvement in basic semantic understanding.
At the same time, the improvement of D-LADAN compared to baselines in terms of law article and charge prediction sub-tasks decreased with the increase of the backbone's capacity.
This indicates that the pre-trained model has a certain ability to distinguish subtle differences.}

\item{From Tables~\ref{tab:CAIL-small} and~\ref{tab:CAIL-big}, we see that the performance of Few-Shot on charge prediction is competitive, but its performance in the terms of penalty prediction is far from ideal. This is because the ten predefined attributes of Few-Shot are only effective for identifying charges instead of the term of penalty. This comparison also proves the robustness of our D-LADAN.}

\item{While other multi-task methods (i.e., TOPJUDGE, MPBNF-WCA, R-former, and NeurJudge) degenerate into base models (cf. Table~\ref{tab:Criminal-all}), the outstanding performance of D-LADAN to solve the single sub-task of charge prediction on the Criminal datasets prove its powerful generalization ability or transferability.}

\item{For datasets of each series (i.e., the CAIL datasets and the Criminal datasets), larger ones always yield better performance than smaller ones.
Such a result also conforms to the intuition that more abundant samples can make the model learn more accurate knowledge.}

\end{enumerate}

\begin{figure*}[t]
  \centering
  \includegraphics[width=1.0\linewidth]{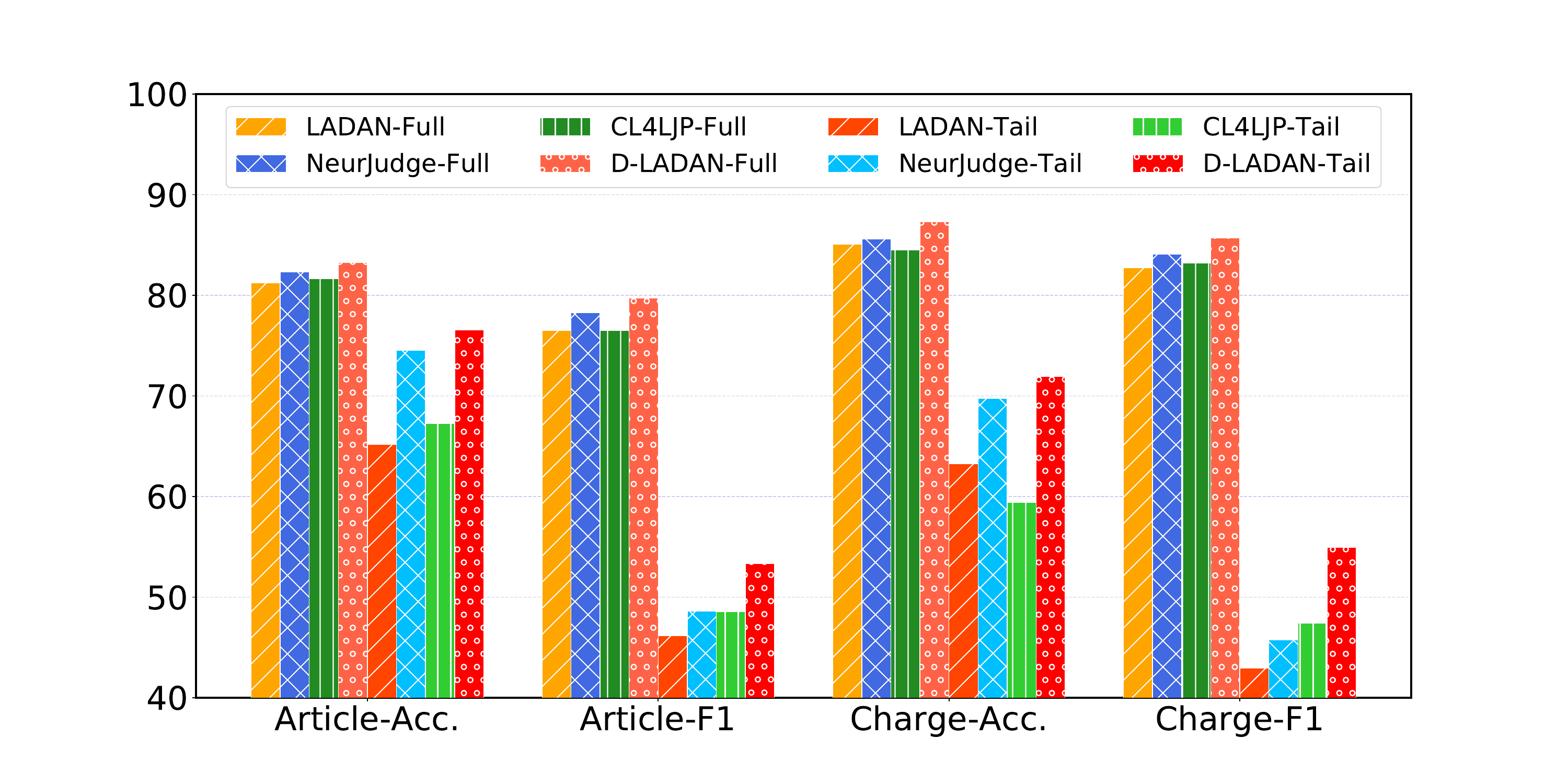}
\caption{Performance on the tail categories of the CAIL-small dataset. We compare the D-LADAN with LADAN, Neurjudge, and CL4LJP. To show that the improvement in the tail category is significant, we also provide performance on the full categories.}
\label{fig:CAIL_Tail}
\end{figure*}

\subsection{Study of Data Imbalance}\label{subsec: Imbalance Expreiment}
To verify the advance of our model in dealing with the imbalance problems, we have constructed a rich variety of experiments, including:

\begin{enumerate}
\item{
In view of the phenomenon shown in Fig.~\ref{fig:Acc_frequency}, to verify that our D-LADAN can effectively improve the performance of tail categories, we investigate D-LADAN's performance under this circumstance from the perspective of instances.
Referring to the setup of CL4LJP~\cite{zhang2023CL4LJP}, we test D-LADAN on the cases of tail law articles and charges, which contain fewer than 200 cases in the CAIL-small dataset. 
As for the term of penalty, we do not test on this sub-task due to its relatively balanced distribution of class labels, i.e., each class has more than 200 samples. 
As shown in Fig.~\ref{fig:CAIL_Tail}, although all models perform worse on the tail classes of the CAIL-small dataset than in all categories, D-LADAN improves performance more on the tail classes over the other three baselines, especially on the F1-score metric. 
This experimental result proves that the improvement of D-LADAN for the tail categories is significant. Notice that LADAN's tail performance is lower than the baselines, i.e., Neurjudge and CL4LJP, we can assert that the significant improvement in tail categories comes from our proposed revised memory mechanism for D-LADAN. By dynamically sensing the posterior semantic similarity relationships between law articles (and charges), D-LADAN can learn how to adaptively revise the inductive bias caused by the data imbalance problem.
}

\begin{table}[t]
\centering
\caption{
Macro-F1 values of various charges on Criminal-S. As this experiment is proposed by the Few-shot baseline, the improvement in parentheses, i.e., $(\uparrow)$, is also relative to the Few-shot baseline. Besides, the results of our D-LADAN are in \textbf{bold}.
}\label{tab:imbalance problem}
\begin{tabular}{lccc}
\toprule
\bf{Frequency} & Low & Medium & High \\ 
\midrule
\bf{Number} & 49 & 51 & 49 \\ 
\midrule

CNN             & $ 15.20 $ & $ 45.78 $ & $ 78.02 $  \\
HARNN           & $ 32.64 $ & $ 55.02 $ & $ 83.28 $  \\
FLA             & $ 26.31 $ & $ 51.42 $ & $ 82.60 $  \\
Few-shot        & $ \underline{49.71} $ & $ \underline{60.02} $ & $ \underline{85.23} $  \\
GFDN            & $ 52.11 $ & $ 67.80 $ & $ 88.90 $  \\
NeurJudge      & $ 48.98 $ & $ 59.17 $ & $ 87.90 $  \\
CL4LJP          & $ 34.69 $ & $ 48.47 $ & $ 84.28 $  \\
\midrule
LADAN           & $ 51.02(\uparrow \underline{1.31\%}) $ 
                & $ 67.10(\uparrow \underline{7.08\%}) $ 
                & $ 88.82(\uparrow \underline{3.59\%}) $  \\
                
\bf D-LADAN     & $\bf 59.18(\uparrow \underline{9.33\%}) $ 
                & $\bf 69.37(\uparrow \underline{9.35\%}) $ 
                & $\bf 89.11(\uparrow \underline{3.88\%}) $  \\  
\bottomrule
\end{tabular}%
\end{table}

\item{
To further verify that our method effectively solves the data imbalance problem, we also construct a comparative experiment from the perspective of categories.
Following the experimental setup of~\cite{hu2018few}, we restrict the situation to the charge prediction task to avoid interference from a multi-task framework and evaluate the performance of D-LADAN on charges with different frequencies.
Table~\ref{tab:imbalance problem} shows the results, where improvement with underline is relative to Few-shot.
We see that D-LADAN improves the prediction accuracy of low-frequency (i.e., few-shot) charges most significantly, up to $9.33\%$, and outperforms all baselines in all evaluation indexes. This result demonstrates the effectiveness of D-LADAN in solving the data imbalance problem.
}

\begin{figure*}[t]
  \centering
  \subfigure[\label{fig:Law_Gap}]{\includegraphics[width=0.95\linewidth]{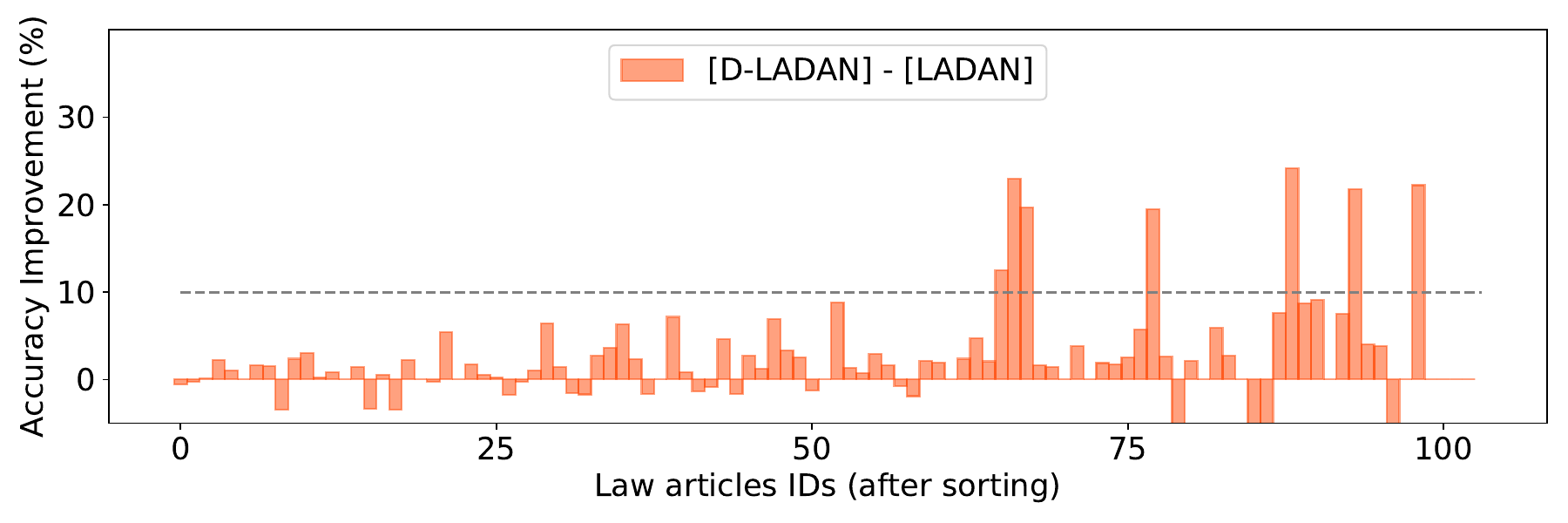}} 
  \subfigure[\label{fig:Accu_Gap}]{\includegraphics[width=0.95\linewidth]{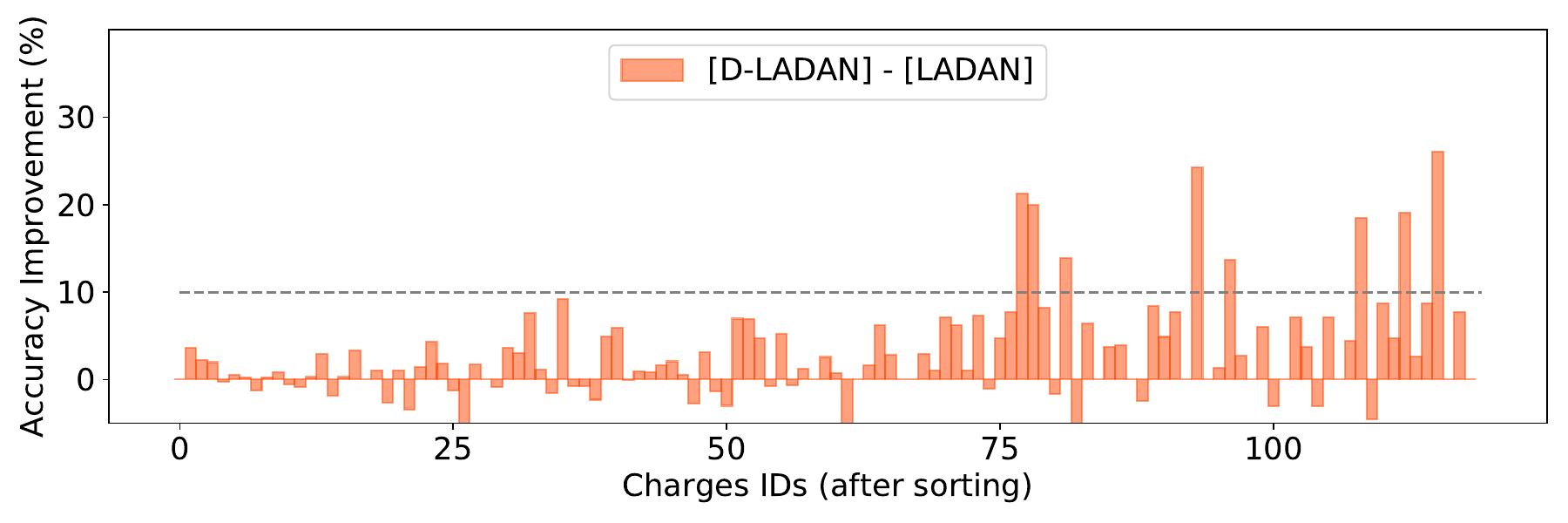}}
\caption{Accuracy improvements of D-LADAN compared with LADAN on law articles and charges with different frequencies. Note that the IDs of the X-axis have been sorted in descending order of frequency, and the dataset used is CAIL-small.}
\label{fig:Acc_Gap}
\end{figure*}

\item{
To prove the remarkable contribution of the momentum-updated revised memory mechanism in solving the imbalance problem, we follow the principle of controlling variables and make a fine-grained comparison between D-LADAN and LADAN on the CAIL-small datasets, which also echoes Fig.~\ref{fig:Acc_frequency}. According to the results shown in Fig.~\ref{fig:Acc_Gap}, for both law article prediction and charge prediction, the categories with a significant improvement of over $10\%$ concentrate in the tail (i.e., few-shot classes).
Such results indicate that the revised memory mechanism effectively corrects the negative bias caused by the imbalanced problem and enables the model to correctly identify the differences between classes.
}
\end{enumerate}

\subsection{Ablation Experiments}
To further illustrate the significance of considering the difference between law articles, we conducted ablation experiments on model \textit{D-LADAN+MTL} with dataset CAIL-small. The ablation variations include:
\begin{itemize}
\item {\bf -no RM}: To show the importance of our revised memory mechanism (\textbf{RM}), we build a variation model without the RM, i.e., use only the $\tilde{\mathbf{v}}_f = [\mathbf{v}_f^{\text{b}} \oplus \mathbf{v}_f^{\text{p}}]$ to predict all results.

\item {\bf -no GCL}: To evaluate the effectiveness of our graph construction layer (\textbf{GCL}), we build a D-LADAN model with the GCL's removing threshold $\theta=0$, i.e., directly applies the GDO on the fully-connected graph $G^*$ to generate a global distinction vector $\hat{\beta}$ for re-encoding the fact description.

\item {\bf -no GDO}: To verify the effectiveness of our graph distillation operator (\textbf{GDO}), we build a no-GDO D-LADAN model. 
For the constructed prior graph of law articles $G$, we directly pool each prior subgraph $g_i$ to a distinction vector $\beta_i$ without using GDOs. 
And for the posterior graph of revised memory $G_{\mathcal{M}}$, we also discard the weighted GDOs and directly use the original memory as the final revised distinction vector, i.e., $\mathbf{\gamma}_{i} = \mathbf{m}_{L_i}^{(0)} = \mathbf{m}_{L_i}$.

\item {\bf -no All}: To evaluate the importance of considering the difference among law articles, we remove all RM, GCL, and GDO from D-LADAN by setting $\theta=1.0$, i.e., each law article independently extracts the attentive feature from fact description.
\end{itemize}

\begin{table}[t]
\centering
\caption{Ablation analysis on CAIL-small.}\label{tab:ablation_exp}
\begin{tabular}{@{}lcccccc@{}}
\toprule
Tasks & \multicolumn{2}{c}{Law} & \multicolumn{2}{c}{Charge} & \multicolumn{2}{c}{Penalty} \\ \cmidrule(lr){2-3} \cmidrule(lr){4-5} \cmidrule(lr){6-7}
Metrics & \multicolumn{1}{c}{Acc.} & \multicolumn{1}{c}{F1} & \multicolumn{1}{c}{Acc.} & \multicolumn{1}{c}{F1} & \multicolumn{1}{c}{Acc.} & \multicolumn{1}{c}{F1} \\ \midrule
D-LADAN+MTL     & $ \bf 83.24 $ & $ \bf 79.73 $
                & $ \bf 86.95 $ & $ \bf 85.48 $
                & $ \bf 40.97 $ & $ \bf 36.89 $ \\
-no RM          & $ 81.20 $ & $ 76.47 $
                & $ 85.07 $ & $ 82.74 $
                & $ 38.29 $ & $ 32.65 $ \\
-no GCL         & $ 81.97 $ & $ 78.25 $
                & $ 86.10 $ & $ 84.83 $
                & $ 39.19 $ & $ 35.36 $ \\
-no GDO         & $ 82.20 $ & $ 78.57 $
                & $ 86.30 $ & $ 84.96 $
                & $ 39.46 $ & $ 35.66 $ \\
-no All         & $ 79.79 $ & $ 74.90 $
                & $ 83.80 $ & $ 82.12 $
                & $ 36.17 $ & $ 31.40 $ \\
\bottomrule
\end{tabular}%
\end{table}

Table~\ref{tab:ablation_exp} shows the experimental results.
We see that all RM, GCL, and GDO effectively improve the performance of D-LADAN.
We summarize specific observations as follows:
\begin{enumerate}
\item{The effect of RM is the most significant, and its improvement is up to 4.24\% on the F1-score. This result reflects the importance of revising the negative bias that the model learned, especially for the datasets with imbalanced distributions (cf. Table~\ref{tab:imbalance problem}).}

\item{Compare with {\bf D-LADAN+MTL}, the average F1-score degradation of {\bf -no GCL} is 1.1\%, which is greater than the 0.85\% of {\bf -no GDO}.
This result indicates that the GCL is more critical than the GDO. It's because GDO has a limited performance when the law article communities obtained by GCL are not accurate.}

\item{When removing all RM, GCL, and GDO, the accuracy of D-LADAN decreases to that of HARNN+MTL. This result powerfully demonstrates the effectiveness of our method in exploiting differences between similar law articles.}
\end{enumerate}  

\begin{figure*}[htp]
\centering
  \subfigure[\label{fig:case_185}]{\includegraphics[width=1.0\linewidth]{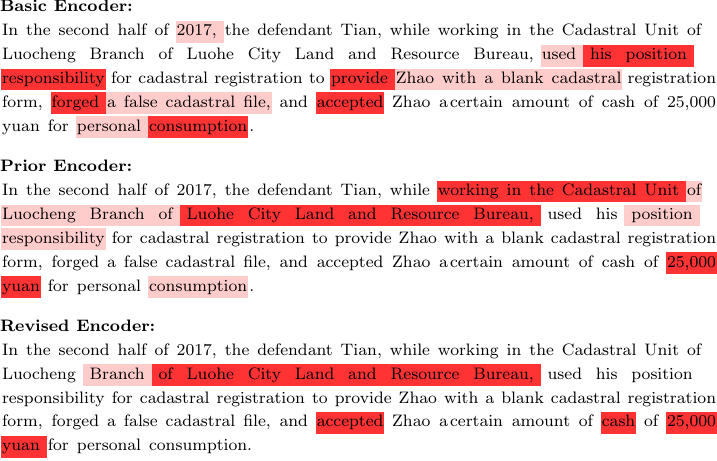}}
  \subfigure[\label{fig:case_163}]{\includegraphics[width=1.0\linewidth]{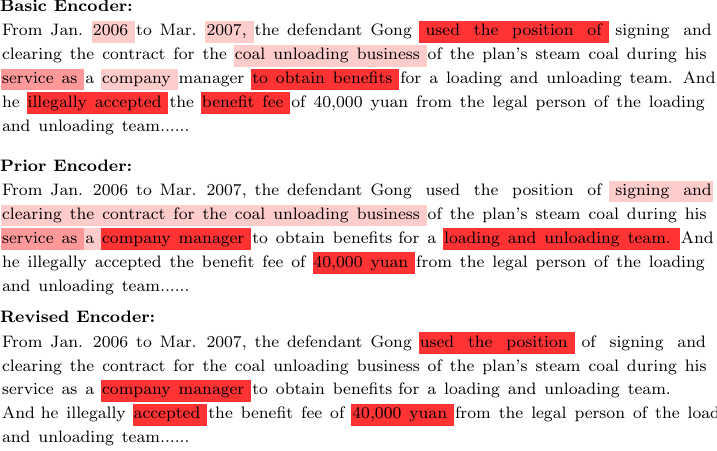}}
\caption{The attention visualization on case examples. \textbf{(a)} Case example of Law Article 185: Crime of acceptance of bribes; \textbf{(b)} Case example of Law Article 163: Bribery crime of non-state employees.}
\label{fig:case_study}
\end{figure*}

\subsection{Case Study} \label{sec:case_study}
To intuitively verify that D-LADAN effectively extracts distinguishable features, we visualize the attention mechanism of D-LADAN's encoders.
Fig.~\ref{fig:case_study} shows two law case examples that correspond to \textit{Article 385} and \textit{Article 163} respectively, where the darker the word is, the higher the attention score it gets in the corresponding encoder, i.e., its information is more important to the corresponding encoder.
For the basic encoder, we see that the vital information in these two cases is very similar, in 
 which both contain the word like \textit{``use position''}, \textit{``accept benefit''}, \textit{``accept ... cash''}, etc. 
 Therefore, when using just the representation of the basic encoder to predict acceptable law articles, charges, and terms of penalty, these two cases tend to be misjudged.
As we mentioned in Section~\ref{sec:re_encoder}, with the prior distinction vector, our prior encoder focuses on extracting distinguishable features like defendants' identity information (e.g., \textit{``company manager'' and ``working in the Cadastral Unit of Luocheng Branch of Luohe City Land and Resources Bureau''} in our examples), which partly distinguish the applicable law articles and charges of these two cases.
In addition, we notice that the revised encoder focuses on the valuable information more purely, where the words with light color are almost absent. This is because the revised memory mechanism needs to reinforce further the valuable information for overcoming the imbalance problem, thus it deprecates all the unimportant parts of the fact description.

\subsection{Robustness of Different Backbones} \label{sec:different_backbones}
As the framework of D-LADAN is backbone-independent, it is easy for it to transfer across different backbone models.
In this section, we take an experiment to demonstrate that the D-LADAN framework can deliver improvements on different backbone models.
We choose three different backbone models, in addition to the HARNN and BERT used in the previous experiments, we also compare another:
\begin{itemize}
\item {\bf Lawformer~\cite{xiao2021lawformer}}: a PLM model with the RoBERTa structure, which is obtained by fine-tuning on Chinese legal long documents using roberta-wm-ext \cite{cui2021pre} checkpoint. 
\end{itemize}
Since Lawformer can accept longer input lengths, we set the maximum document length as $1,024$ for it.

As shown in Table~\ref{tab: Different-backbones}, our D-LADAN framework achieves significant improvements on all backbone models, it further proves the effectiveness of the framework.
In addition, we notice that the absolute boost obtained by the DLADAN framework tends to become smaller for more capable backbone models.
This may reflect the fact that more linguistically competent models have themselves mastered some ability to distinguish subtle differences.

\begin{table*}[t]
\centering
\small
\caption{
Judgment prediction results on CAIL-small with different backbones. The results of our D-LADAN are in \textbf{bold}. $\dagger$ denotes D-LADAN achieves significant improvements over all existing baselines in paired t-test with $p$-value $< 0.05$.
}\label{tab: Different-backbones}
\resizebox{\textwidth}{!}{%
\begin{tabular}{lcccccccccccc}
\toprule
Tasks              & \multicolumn{4}{c}{Law Articles}                     & \multicolumn{4}{c}{Charges}                          & \multicolumn{4}{c}{Term of Penalty}                   \\ 
 \cmidrule(lr){2-5} \cmidrule(lr){6-9} \cmidrule(lr){10-13}
Metrics            & Acc.        & MP          & MR          & F1
                    & Acc.        & MP          & MR          & F1
                    & Acc.        & MP          & MR          & F1          \\
 \midrule

HARNN          
                    & $79.79$     & $75.26$     & $76.79$     & $74.90$
                    & $83.80$     & $82.44$     & $82.78$     & $82.12$
                    & $36.17$     & $34.66$     & $31.26$     & $31.40$     \\

\bf{D-LADAN}    
                    & $ \bf 83.24^{\dagger} $ & $ \bf 80.32^{\dagger} $ 
                    & $ \bf 81.46^{\dagger} $ & $ \bf 79.73^{\dagger} $
                    & $ \bf 86.95^{\dagger} $ & $ \bf 85.69^{\dagger} $ 
                    & $ \bf 86.17^{\dagger} $ & $ \bf 85.48^{\dagger} $
                    & $ \bf 40.97^{\dagger} $ & $ \bf 38.31^{\dagger} $ 
                    & $ \bf 37.06^{\dagger} $ & $ \bf 36.89^{\dagger} $ \\

\midrule
BERT    
                    & $ 83.73 $     & $ 80.87 $     & $ 81.75 $     & $ 80.10 $
                    & $ 87.48 $     & $ 86.21 $     & $ 85.75 $     & $ 85.69 $
                    & $ 41.16 $     & $ 39.91 $     & $ 38.08 $     & $ 38.37 $     \\
\bf $\text{D-LADAN}_{BERT}$    
                    & $ \bf 85.88^{\dagger} $ & $ \bf 84.56^{\dagger} $ 
                    & $ \bf 83.52^{\dagger} $ & $ \bf 82.38^{\dagger} $
                    & $ \bf 89.93^{\dagger} $ & $ \bf 88.78^{\dagger} $ 
                    & $ \bf 88.79^{\dagger} $ & $ \bf 88.48^{\dagger} $
                    & $ \bf 43.33^{\dagger} $ & $ \bf 42.74^{\dagger} $ 
                    & $ \bf 40.07^{\dagger} $ & $ \bf 40.90^{\dagger} $ \\
\midrule
Lawformer         & $ 84.39 $     & $ 81.56 $     
                    & $ 82.76 $     & $ 81.02 $
                    & $ 89.65 $     & $ 88.44 $     
                    & $ 88.73 $     & $ 88.27 $
                    & $ 43.05 $     & $ 42.76 $     
                    & $ 40.10 $     & $ 41.23 $     \\
\bf $\text{D-LADAN}_{Lawformer}$   
                    & $ \bf 86.09^{\dagger} $ & $ \bf 85.21^{\dagger} $ 
                    & $ \bf 84.05^{\dagger} $ & $ \bf 83.40^{\dagger} $
                    & $ \bf 90.16^{\dagger} $ & $ \bf 89.81^{\dagger} $
                    & $ \bf 89.23^{\dagger} $ & $ \bf 89.05^{\dagger} $
                    & $ \bf 44.36^{\dagger} $ & $ \bf 44.07^{\dagger} $
                    & $ \bf 40.96^{\dagger} $ & $ \bf 42.12^{\dagger} $ \\
\bottomrule
\end{tabular}%
}
\end{table*}

\subsection{Optimal Weight of Training Loss}\label{sec:loss_weight}
To explore the influence of the prior community selection and revised memory selection, we experiment with groups of different hyper-parameters $(\lambda_{c}, \lambda_{m})$ of loss function on the CAIL-small dataset to find the optimal combination.
In this experiment, we use the grid search strategy and set the search range of each hyper-parameter to $[0.1, 0.25, 0.5, 0.75, 1.0]$
\footnote{Since the experimental results in Section 5.5 demonstrate that the absence of either $\mathscr{L}_{c}$ or $\mathscr{L}_{m}$ significantly damages performance, we do not start with $0$ for hyper-parametric search.} by using the average F1 score of the three sub-tasks as the evaluation metric.

\begin{figure*}[t]
\centering
 \subfigure[\label{fig: optimal_avgF1}]{\includegraphics[width=0.246\linewidth]{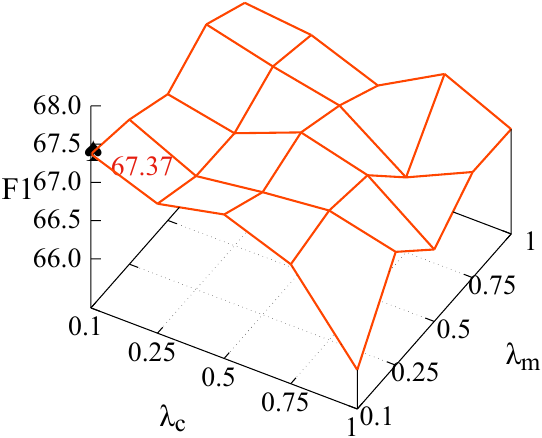}}
 \subfigure[\label{fig: optimal_law}]{\includegraphics[width=0.246\linewidth]{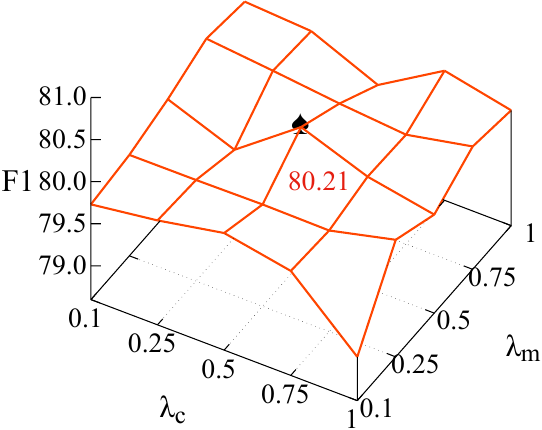}}
 \subfigure[\label{fig: optimal_charge}]{\includegraphics[width=0.24\linewidth]{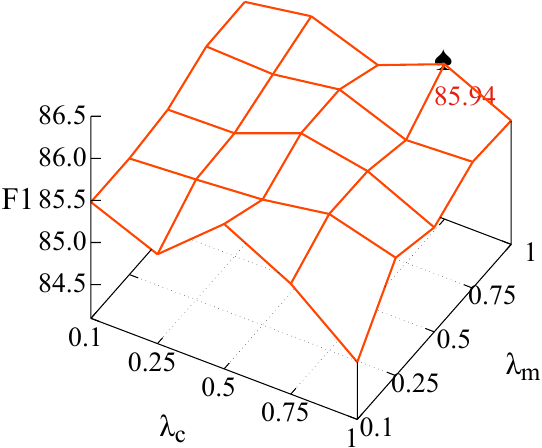}}
 \subfigure[\label{fig: optimal_time}]{\includegraphics[width=0.246\linewidth]{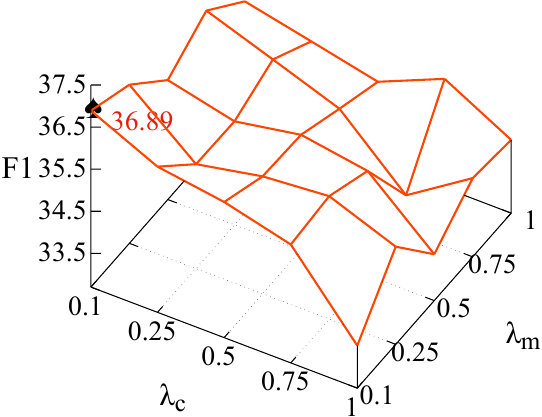}}
\caption{The grid search of optimal weight based on different evaluation metrics is carried out on the CAIL-small dataset:
(a) The average F1 score of three sub-tasks;
(b) The F1 score of the law article prediction task;
(c) The F1 score of the charge prediction task;
(d) The F1 score of the term of penalty  prediction task;
}
\label{fig: optimal_weight}
\end{figure*}

The results are shown in Fig.~\ref{fig: optimal_weight}, from the overall performance point of view, the optimal weights in the loss function are $(0.1, 0.1)$.
At the same time, we show the results of the sub-tasks respectively (as shown in Fig.~\ref{fig: optimal_law},~\ref{fig: optimal_charge}, and~\ref{fig: optimal_time}), and obtain the following interesting experimental observations,
\begin{enumerate}
\item{
For the law prediction sub-task, its experimental performance remains stable across various settings of the weight hyper-parameter.
This stability results in minimal fluctuations, barring a few outliers.
This is because this task benefits both from using the text definition of law articles as prior knowledge and from updating the revised memory $\mathcal{M}$ with the weights of the law article classifier.
}

\item{
For the charge prediction sub-task, the experimental performance exhibits robustness to hyper-parameter $\lambda_{c}$ while experiencing slight enhancements with increasing values of hyper-parameter $\lambda_{m}$.
This is because the charge prediction task mainly benefited from the corresponding correction revised memory $\mathcal{M}_{c}$ when this task can only obtain limited help from prior knowledge through the potential relationship between law articles and charges.
}

\item{
As for the term of penalty prediction sub-task, we find that whether the hyper-parameter $\lambda_{c}$ or $\lambda_{m}$ increases, the experimental performance has a significant downward trend.
This is because the help obtained by this sub-task from both prior knowledge and revised memories is latent.
When the model focuses too much on such less related prior and posterior knowledge, it will ignore the main information that the penalty prediction sub-task focuses on and harm the experimental results.
}
\item{
We also focus on an outlier $(1,0.1)$ where all three tasks achieve the worst results under this hyper-parameter setting. 
This may indicate that when the ratio of $\lambda_{c}: \lambda_{m}$ is too large, the model would ignore the posterior knowledge of revised memory.
}
\end{enumerate}

\section{Discussion: D-LADAN V.S. LLMs} \label{sec: discussion}
In the last year or two, the proposed large language models (LLMs) represented by GPT-4, have turned the landscape of NLP upside down. Due to the excellent performance of large models, it has reached the human level in many simple NLP tasks and even surpassed it in some tasks.
Therefore, we deliberately added this section to explore the practice of the new paradigm of LLMs in the task of LJP, and to further reflect the application value of the work in this paper by comparing the prediction performance of our D-LADAN and LLMs.

In this section, we select four state-of-the-art open-source general LLMs (i.e., \emph{Llama-2-7b-128k}\footnote{\url{https://huggingface.co/NousResearch/Yarn-Llama-2-7b-128k}}, \emph{Qwen1.5-7B-Chat}\footnote{\url{https://huggingface.co/Qwen/Qwen1.5-7B-Chat}}, \emph{Yi-6B-200k}\footnote{\url{https://huggingface.co/01-ai/Yi-6B-200K}}, and \emph{chatglm3-6b-32k}\footnote{\url{https://huggingface.co/THUDM/chatglm3-6b-32k}}) and two closed-source LLMs (i.e., \emph{GPT-3.5}\footnote{Model name: \textit{gpt-3.5-turbo-1106}} and \emph{GPT-4}\footnote{Model name: \textit{gpt-4-turbo-preview}}) for evaluation.
 
For the charge prediction task, since the charge name is short enough to meet the input length limit of LLMs and has certain semantics, we conduct zero-shot experiments on LLMs using the prompts for text classification. 
The reason for not providing some instances like the in-context learning setting is that the fact description of each case is too long and the input length limit of LLMs cannot provide the corresponding instance for each charge.
Here, we provide the prompt used to make the charge prediction in the following, where the black part represents the fixed prompt texts, and the gray part shows input and output information for the prediction of an actual case, including the set of charges, the fact description, and the predicted result of model,
\vspace{4pt}
\begin{mdframed}[linewidth=1pt, linecolor=black, frametitle={Instruction for Charge Prediction Task}]
\begin{quote}
    Instruction: The list of all candidate charges includes: {\color{gray} Crime of obstructing public service, Crime of picking quarrels and provoking trouble, \dots }. Please determine the charges according to the following description of the case. Note that only the most relevant charge is printed.\\
    User: Description of the case: {\color{gray} At about 20 o'clock on March 28, 2016, the defendant Yan picked up a VIVOX 5 mobile phone belonging to the victim Xie on the roadside of the football stadium in Mahu New Village, Hongshan District of the city, and from 21 o 'clock on March 28 of the same year, he secretly stole RMB 3,723 yuan from the victim Xie's Alipay payment through the payment of small amounts without secret payment function several times \dots }\\
    Assistant: The most applicable charge is {\color{gray} Crime of larceny.}
\end{quote}
\end{mdframed}
\vspace{4pt}
For the law article prediction task, since the serial numbers of law articles have no semantic meaning, we choose the text definition of law articles as the category labels.
However, due to the LLM input length limit, we cannot add text definitions of all laws to the prompt.
Thus, to solve the law article prediction task, we adopted the strategy of Retrieval-Augmented Generation (RAG)~\cite{lewis2020retrieval}.
In practice, we first use BM25~\cite{robertson1994some} to select the top $10$ relevant law articles from candidates.
Then, we fill these ten law articles and the fact description into our prompt.
Finally, the synthesized prompt was input into LLMs to obtain the final prediction result.
In the following, we also show our prompt for the law article prediction task,  where the black part represents the fixed prompt texts, and the gray part shows input and output information for the prediction of an actual case, including the fact description, the set of selected law articles, and the predicted result of model,

\vspace{4pt}
\begin{mdframed}[linewidth=1pt, linecolor=black, frametitle={Instruction for Law Prediction Task}]
\begin{quote}
    Instruction: Please determine which law the current case violates based on the description of the case and the candidate law provided. Note that only output the most applicable law article.\\
    User: Description of the case: {\color{gray} At about 20 o'clock on March 28, 2016, the defendant Yan picked up a VIVOX 5 mobile phone belonging to the victim Xie on the roadside of the football stadium in Mahu New Village, Hongshan District of the city, and from 21 o 'clock on March 28 of the same year, he secretly stole RMB 3,723 yuan from the victim Xie's Alipay payment through the payment of small amounts without secret payment function several times \dots }\\
    \indent \quad \quad \ \  Candidate law articles: {\color{gray} Article 236: Whoever rapes a woman by violence \dots; Article 237: Whoever forcibly indecently assaults or humiliates a woman by violence, coercion or other means, \dots; }\\
    Assistant: The most applicable law article is {\color{gray} Article 264 Whoever steals public or private property if the amount involved is relatively large, or commits repeated theft, housebreaking theft, theft with murder weapon or pickpocketing, shall be sentenced to fixed-term imprisonment of not more than three years, criminal detention or public surveillance and shall also, or shall only, be fined \dots}
\end{quote}
\end{mdframed}
\vspace{4pt}

We conduct comparative experiments on the CAIL-small dataset, and Table~\ref{tab:LLMs} shows the result.
In a word, the natural language generation (NLG) based LLMs are unsatisfactory in solving the neural language understanding (NLU) based LJP problems.
We summarize the detailed experimental analysis as follows,
\begin{enumerate}
\item{
Among all LMMs, \emph{GPT-4} performs the best, significantly better than the other LLMs due to its larger model size and better engineering.
}

\item{
The \emph{Llama-2-7b-128k} model obtains the worst experimental results. 
After our instance-level analysis, it is found that \emph{Llama-2-7b-128k} often cannot understand Chinese prompts and fact descriptions because its pre-training corpus is all in English.
}

\end{enumerate} 

Because of the problems encountered in the implementation process, we believe that LLMs can be explored from the following aspects in solving the LJP task,

\begin{enumerate}
\item{
\textbf{Dedicated LLMs}. Using the corpus in the judicial domain to retrain or continue training a dedicated LLM can help it understand the professional knowledge in the judicial domain.
}

\item{
\textbf{A new paradigm for the LJP task}. Although LLMs perform well in solving some classification tasks with few categories as present, they perform poorly in classification tasks with many categories such as LJP, and even some extreme classification tasks.
So how to represent the classification task with many categories into a more generative-model-friendly paradigm is also an idea to solve the LJP problem.
}

\item{
\textbf{Some more efficient process/engineering design}. 
As we use the RAG strategy to improve the performance of the law article prediction task, we conjecture that a retrieval model with higher accuracy can further improve the performance. 
At the same time, we also think that if some summary processes are added to shorten the input length of knowledge so that LLMs can receive more knowledge from the limited input, it may also bring improvement.
In addition, the chain of thought (COT) strategy~\cite{wei2022chain} may also improve performance, such as having the large model first find the key legal information in the fact description, then compare this information with each law/crime, and finally make a prediction.
}
\end{enumerate} 

\begin{table*}[h]
\centering
\caption{
Judgment prediction results of LLMs on CAIL-small.
}\label{tab:LLMs}

\begin{tabular}{lcccccccc}
\toprule
Tasks              & \multicolumn{4}{c}{Law Articles}   & \multicolumn{4}{c}{Charges}  \\ 
 \cmidrule(lr){2-5} \cmidrule(lr){6-9}
Metrics            & Acc.        & MP          & MR          & F1
                    & Acc.        & MP          & MR          & F1          \\
 \midrule
\multicolumn{9}{c}{\textbf{Open Source}} \\
 \midrule
Llama-2-7b-128k    & $13.61 $     & $16.11 $     & $14.15 $     & $9.79 $
                    & $ 14.46 $     & $ 10.00 $     & $ 9.16 $     & $ 7.67 $     \\
                    
Yi-6B-200k         & $ 24.15 $     & $ 33.14 $     & $ 22.59 $     & $ 19.65 $
                    & $ 40.08 $     & $ 31.88 $     & $ 32.56 $     & $ 29.14 $     \\

Qwen1.5-7B-Chat    & $ 46.91 $     & $ 49.28 $     & $ 40.72 $     & $ 36.01 $
                    & $ 34.88 $     & $ 26.84 $     & $ 30.78 $     & $ 24.69 $     \\

chatglm3-6b-32k    & $ 41.98 $     & $ 47.89 $     & $ 36.12 $     & $ 33.59 $
                    & $ 40.86 $     & $ 28.65 $     & $ 31.63 $     & $ 27.41 $     \\

 \midrule
\multicolumn{9}{c}{\textbf{Close Source}} \\  
 \midrule
GPT-3.5            & $ 50.65 $     & $ 51.52 $     & $ 46.92 $     & $ 42.55 $
                    & $ 43.78 $     & $ 34.21 $     & $ 37.01 $     & $ 31.67 $     \\

GPT-4              & $ 61.77 $     & $ 61.22 $     & $ 55.67 $     & $ 51.57 $
                    & $ 60.98 $     & $ 53.71 $     & $ 55.94 $     & $ 49.78 $     \\

 \midrule
D-LADAN          & $ {\bf 83.24} $ & $ {\bf 80.32} $ 
                    & $ {\bf 81.46} $ & $ {\bf 79.73} $
                    & $ \bf 86.95 $ & $ \bf 85.69 $ 
                    & $ \bf 86.17 $ & $ \bf 85.48 $ \\

D-$\text{LADAN}_{BERT}$   
                    & $ \bf 85.88 $ & $ \bf 84.56 $ 
                    & $ \bf 83.52 $ & $ \bf 82.38 $
                    & $ \bf 89.93 $ & $ \bf 88.78 $ 
                    & $ \bf 88.79 $ & $ \bf 88.48 $  \\

D-$\text{LADAN}_{Lawformer}$   
                    & $ \bf 86.09 $ & $ \bf 84.63 $ 
                    & $ \bf 84.05 $ & $ \bf 82.71 $
                    & $ \bf 90.16 $ & $ \bf 89.81 $ 
                    & $ \bf 89.00 $ & $ \bf 88.85 $  \\
\bottomrule
\end{tabular}%
\end{table*}
\section{Conclusion} \label{sec: conclusion}
In this paper, we present an end-to-end model, D-LADAN, to solve both confusing law articles (or charges) and the data imbalance problem in LJP.
We propose an effective attention mechanism to extract the key features for distinguishing confusing law articles attentively.
Our attention mechanism not only considers the interaction between fact description and law articles but also the differences among similar law articles, which are effectively extracted by a novel graph neural network named GDL.
In addition, to solve the imbalance problem, our D-LADAN uses a novel momentum updated memory mechanism to capture the semantic similarity relation of labels that the model learned, which combines with our attention mechanism to revise the model's negative biased understanding of the above relation and further improve the performance.
The experimental results on real-world datasets show that our D-LADAN raises the F1-score of state-of-the-art by up to $2.93$\%, and the accuracy of the few-shot classes by $7.07$\% averagely.
In the future, we plan to study complicated situations such as law cases with multiple defendants and charges.

\begin{acks}
The authors would like to thank all reviewers for their comments and suggestions. 
This work was supported in part by the National Key R\&D Program of China
(2021YFB1715600), National Natural Science Foundation of China (61902305, 61922067), MoE-CMCC "Artifical Intelligence" Project (MCM20190701). A preliminary version of this work was published as a long paper at the 58th Annual Meeting of the Association for Computational Linguistics (ACL 2020).

\end{acks}
\bibliographystyle{ACM-Reference-Format}
\bibliography{acmart,acl2020}










\end{document}